\documentclass[twoside,journey]{IEEEtran}
\usepackage{makecell}
\usepackage{hyperref}
\usepackage{array}
\usepackage{caption2}
\usepackage{graphicx,amssymb,amsmath}
\usepackage{multicol}
\usepackage[noadjust]{cite}
\usepackage{setspace}
\usepackage{subfigure}
\usepackage{graphicx}
\usepackage{float}
\usepackage{url}
\usepackage{stfloats}
\usepackage{amsthm,pifont}
\usepackage{flushend}
\usepackage{cases,subeqnarray}
\usepackage{bm,multirow,bigstrut}
\usepackage{amsmath, amsthm, amssymb}
\usepackage{textcomp}
\usepackage{latexsym,bm}
\usepackage{booktabs}
\usepackage{xcolor}
\usepackage{mathtools}
\usepackage{dsfont}
\usepackage{extarrows}
\usepackage{epsfig}
\usepackage{epsfig}
\usepackage{epstopdf}
\usepackage[noend]{algpseudocode}
\usepackage{algorithmicx,algorithm}
\usepackage{calc}
\theoremstyle{plain}

\theoremstyle{plain}

\usepackage{amsmath}

\usepackage{graphicx}
\usepackage{diagbox}
\usepackage{enumitem}
\IEEEoverridecommandlockouts
\begin{document}
%----------------------------title&author&thanks----------------------------
\title{Generative AI for Unmanned Vehicle Swarms: Challenges, Applications and Opportunities}

\author{Guangyuan Liu,  Nguyen Van Huynh,  Hongyang Du, Dinh Thai Hoang, Dusit Niyato,~\IEEEmembership{Fellow,~IEEE}, Kun Zhu,  Jiawen~Kang, Zehui~Xiong, Abbas Jamalipour,~\IEEEmembership{Fellow,~IEEE}, and Dong In Kim,~\IEEEmembership{Fellow,~IEEE}
\thanks{G.~Liu, and H.~Du are with the School of Computer Science and Engineering, the Energy Research Institute @ NTU, Interdisciplinary Graduate Program, Nanyang
Technological University, Singapore (e-mail: liug0022@e.ntu.edu.sg, hongyang001@e.ntu.edu.sg).}
\thanks{N. V. Huynh is with the Department of Electrical Engineering and Electronics, University of Liverpool, Liverpool, L69 3GJ, United Kingdom (e-mail: huynh.nguyen@liverpool.ac.uk)}
\thanks{D. T. Hoang is with the School of Electrical and Data Engineering, University of Technology Sydney, Australia (e-mail: Hoang.Dinh@uts.edu.au)
}
\thanks{D. Niyato is with the School of Computer Science and Engineering, Nanyang Technological University, Singapore (e-mail: dniyato@ntu.edu.sg).}
 \thanks{K. Zhu is with the College of Computer Science and Technology, Nanjing University of Aeronautics and Astronautics, Nanjing, China. (E-mail: zhukun@nuaa.edu.cn).}
\thanks{J. Kang is with the School of Automation, Guangdong University of Technology, China. (e-mail: kavinkang@gdut.edu.cn).}
\thanks{Z. Xiong is with the Pillar of Information Systems Technology and Design, Singapore University of Technology and Design, Singapore (e-mail:
zehui\_xiong@sutd.edu.sg).}
\thanks{A. Jamalipour is with the School of Electrical and Information Engineering, University of Sydney, Australia (e-mail: a.jamalipour@ieee.org).}
\thanks{D. I. Kim is with the College of Information and Communication Engineering, Sungkyunkwan University, South Korea (e-mail: dikim@skku.ac.kr).}
 }
\maketitle

%----------------------------abstract----------------------------
\vspace{-1cm}
\begin{abstract}
With recent advances in artificial intelligence (AI) and robotics, unmanned vehicle swarms have received great attention from both academia and industry due to their potential to provide services that are difficult and dangerous to perform by humans. However, learning and coordinating movements and actions for a large number of unmanned vehicles in complex and dynamic environments introduce significant challenges to conventional AI methods. Generative AI (GAI), with its capabilities in complex data feature extraction, transformation, and enhancement, offers great potential in solving these challenges of unmanned vehicle swarms. For that, this paper aims to provide a comprehensive survey on applications, challenges, and opportunities of GAI in unmanned vehicle swarms. Specifically, we first present an overview of unmanned vehicles and unmanned vehicle swarms as well as their use cases and existing issues. Then, an in-depth background of various GAI techniques together with their capabilities in enhancing unmanned vehicle swarms are provided. After that, we present a comprehensive review on the applications and challenges of GAI in unmanned vehicle swarms with various insights and discussions. Finally, we highlight open issues of GAI in unmanned vehicle swarms and discuss potential research directions.
\end{abstract}

\section{Introduction}
In recent years, Unmanned Vehicles (UVs) have emerged as a disruptive technology, revolutionizing various sectors of daily life with applications spanning from package delivery and civilian Internet of Things (IoT) to military uses~\cite{tan2020unmanned, mcenroe2022survey}. Specifically, UVs refer to vehicles, devices, or machines that can operate with limited or without human intervention, e.g., without a human pilot or crew on board. Thanks to this special property, UVs can be used to perform tasks in challenging or hazardous environments. In general, UVs can be classified into Unmanned Aerial Vehicles (UAVs), Unmanned Ground Vehicles (UGVs), Unmanned Surface Vehicles (USVs), and Unmanned Underwater Vehicles (UUVs). As suggested by their names, each type of UV is designed for particular tasks and environments. For instance, UAVs are widely used for aerial photography and filming, environmental and wildlife monitoring, and surveillance~\cite{motlagh2016low, li2018uav} while UGVs can be used for tasks such as transportation and bomb detection. Differently, USVs and UUVs are used for surface and underwater operations, respectively, including oceanographic data collection, underwater exploration, and submarine surveillance~\cite{liu2016unmanned, neira2021review}.

With recent advances in artificial intelligence (AI) and robotics, the concept of UVs has evolved into a whole new level, namely unmanned vehicle swarms. Essentially, an unmanned vehicle swarm is designed by coordinating a group of UVs, e.g., robots, drones, and other autonomous vehicles, to achieve a common objective~\cite{zhou2020uav, venayagamoorthy2004unmanned}. Practically, each vehicle in a swarm can be equipped with its own sensor, processor, and communication capability. To make them efficiently collaborate together, advanced technologies in AI and robotics have been adopted to coordinate their behaviors and perform complex tasks such as autonomous navigation, self-organization, and failure management~\cite{zhou2020uav, puente2022review}. As a result, unmanned vehicle swarms possess various advantages compared to conventional UVs. In particular, they offer scalability and flexibility in operations by dynamically adjusting the number of vehicles depending on specific missions and requirements. Moreover, if several UVs fail to operate in a swarm, the remaining UVs still can work together to ensure the success of their mission. This is particularly useful in missions requiring high levels of resilience and robustness. Finally, by allowing UVs to learn from and collaborate with each other, unmanned vehicle swarms can enable swarm intelligence, as known as collective intelligence~\cite{kennedy2006swarm, giacomossi2021autonomous}, that greatly improves operational efficiency and reliability.

Although playing an important role in unmanned vehicle swarms, conventional AI techniques still face a number of challenges. Particularly, these techniques require a large volume of labeled training data and can only obtain good performance under specific settings. As such, they are extremely vulnerable to the dynamics and uncertainty of the environment which are particularly characteristics of unmanned vehicle swarms, e.g.,  dynamic connections between unmanned vehicles, effects of winds and ocean currents, and sensor uncertainty and diversity in IoT applications. In addition, traditional AI methods may perform poorly in complex scenarios with a large number of UVs and in challenging environments such as underwater, remote regions, and disaster-affected areas. To overcome these challenges of conventional AI techniques, generative AI (GAI) has been widely adopted in the literature recently due to its groundbreaking abilities in understanding, capturing, and generating the distribution of complex and high-dimensional data. Given the potential of GAI in UV swarms, this paper aims to provide a comprehensive survey on the challenges, applications, and opportunities of GAI in enabling swarm intelligence from various perspectives.

There are a few surveys in the literature focusing on the applications of AI for UVs~\cite{lahmeri2021artificial, kurunathan2023machine, sai2023comprehensive, cheng2023ai}. For example, the authors in~\cite{lahmeri2021artificial} study the applications of conventional AI techniques such as deep learning, deep reinforcement learning, and federated learning in UAV-based networks while the authors in~\cite{kurunathan2023machine} provide a more comprehensive survey on applications of machine learning (ML) in operations and communications of UAVs. Differently, in~\cite{cheng2023ai}, the authors provide a review on AI-enabled UAV optimization methods in IoT networks, focusing on AI for UAV communications, swarm routing and networking, and collision avoidance. Similarly, applications of AI/ML for UAV swarm intelligence are also discussed in~\cite{zhou2020uav}. It is worth noting that the aforementioned surveys and others in the literature mainly focus on UAVs and traditional AI methods. To the best of our knowledge, there is no survey in the literature comprehensively covering the development of GAI for UV swarms. The main contributions of our paper can be summarized as follows.

\begin{itemize}
	\item We provide the fundamentals of UV swarms, including their designs and operations across the aerial, ground, surface, and underwater domains as well as practical use cases.
	
	\item We provide an in-depth overview of common GAI techniques, including generative adversarial networks (GAN), variational autoencoder (VAE), generative diffusion model, transformer, and normalizing flows. The key advantages and challenges of each technique in the context of UV swarms are also highlighted in detail.
	
	\item We comprehensively review applications of GAI to various problems in UV swarms such as state estimation, environmental perception, task/resource allocation, network coverage and peer-to-peer communications, and security and privacy. Through reviewing these GAI applications, we provide insights into how GAI can be applied to address emerging problems in UV swarms.
	
	\item We present essential open issues and future research directions of GAI in UV swarms, including scalability, adaptive GAI, explainable swarm intelligence, security/privacy, and heterogeneous swarm intelligence.
\end{itemize}

\begin{figure}[!]
	\centering
	\includegraphics[scale=0.42]{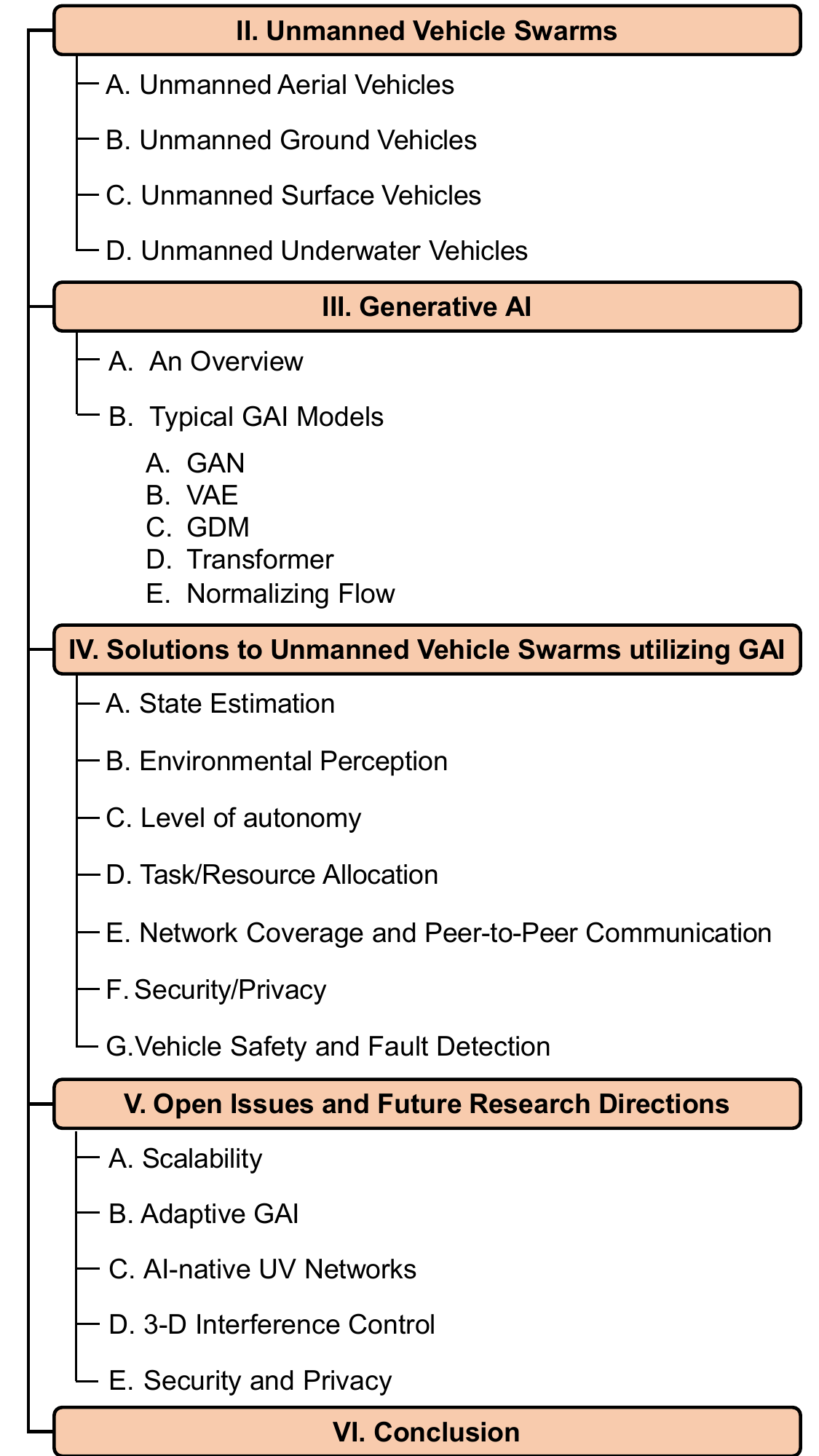}
	\caption{The overall structure of this paper.}
	\label{structure}
\end{figure}

The overall structure of this paper is illustrated in Fig.~\ref{structure}. Section~\ref{USS} provides the fundamentals of UV swarms. An in-depth overview of different GAI techniques and their advantages are presented in Section~\ref{GAI}. Then, Section~\ref{survey} delves into the applications of GAI for emerging problems in UV swarms. Open issues and future research direction of GAI in UV swarms are highlighted in Section~\ref{open_issue}. Section~\ref{conclusion} concludes the paper. Additionally, Table~\ref{tab:abbreviations} lists all the abbreviations used in the paper.

\begin{table*}[!ht]
\centering
\begin{tabular}{llll}
\hline
\textbf{Abbreviation} & \textbf{Description} & \textbf{Abbreviation} & \textbf{Description} \\
\hline
UAV & Unmanned Aerial Vehicle & UGV & Unmanned Ground Vehicle \\
USV & Unmanned Surface Vehicle & UUV & Unmanned Underwater Vehicle \\
VTOL & Vertical Take-Off and Landing & AI & Artificial Intelligence \\
GAI & Generative Artificial Intelligence & GAN & Generative Adversarial Network \\
VAE & Variational Autoencoder & GDM & Generative Diffusion Model \\
NF & Normalizing Flow & DRL & Deep Reinforcement Learning \\
UV & Unmanned Vehicle & LLM & Large Language Model \\
\hline
\end{tabular}
\caption{List of Abbreviations}
\label{tab:abbreviations}
\end{table*}

\section{Unmanned Vehicle swarms}\label{USS}
\begin{figure*}[h!]
\centerline{\includegraphics[width=0.9\textwidth]{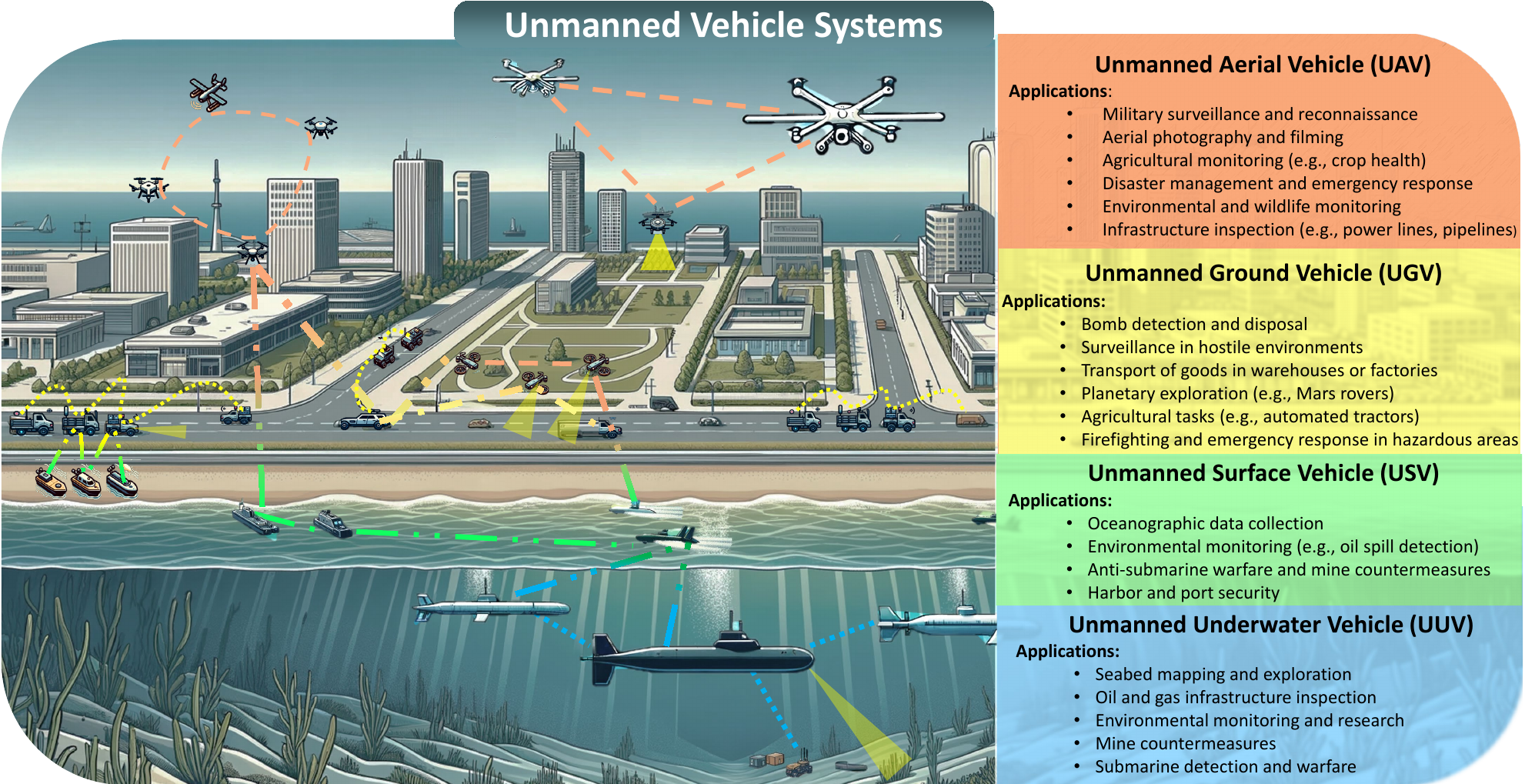}}
\caption{Infrastructure of UV systems and their applications.}
\label{UVapp}
\end{figure*}

 UVs operate across aerial, ground, surface, and underwater domains~\cite{li2023survey} as illustrated in Fig. \ref{UVapp}. They harness advanced technologies like AI and robotics to perform diverse tasks often in challenging or hazardous environments. By forming coordinated groups or swarms, these systems achieve common objectives with enhanced efficiency~\cite{8682048}. UV swarms are categorized primarily based on the degree of autonomy, which ranges from fully autonomous systems, capable of independent operation based on pre-programmed protocols and real-time data, to semi-autonomous systems that require some level of human intervention. UV swarms can also be classified by the hierarchical structure, which includes single-layered or multi-layered configurations, which are integral to the operational dynamics of these swarms. In a multi-layered swarm, leader vehicles are designated to orchestrate the activities of the collective. These leaders communicate with a central server station, constituting the operational hierarchy's peak~\cite{abdelkader2021aerial}. Each UV within the swarm, equipped with substantial computational prowess, is assigned with specific data collection and processing roles. The centralization of mass data processing facilitated either through advanced server stations or cloud-based computing solutions, significantly optimizes the efficiency of data management and task execution.

The diverse applications of UV swarms in various sectors underscore their transformative impact. These applications range from enhancing military operations to revolutionizing agricultural practices, showcasing the blend of agility, precision, and operational efficiency that these systems bring to modern society:

\begin{itemize}

\item \textbf{Surveillance and Monitoring:}
    UVs are widely used for security, survey, monitoring and surveillance purposes. They offer unparalleled efficiency in covering wide areas, reducing manpower requirements, and providing real-time response capabilities, especially in detecting and alerting movements and changes in the environment~\cite{4177711,6045299,9121685,yan2010development}.

    \item \textbf{Environmental Conservation and Management:}
    UVs are actively engaged in environmental monitoring and mapping. They contribute significantly to environmental conservation efforts, including studying ocean biological phenomena, disaster prediction and management~\cite{abdelkader2014optimal,carpentiero2017swarm,s23104643,LIU201671,vamraak2020deep}. Particularly, in events like tsunamis and hurricanes, these systems prove critical, especially in areas rendered inaccessible. They play a vital role in damage assessment, emergency response, and in planning and executing effective disaster management strategies~\cite{qin2016design,hayajneh2016drone}.
    \item \textbf{Entertainment and recreational events:}
    UVs have revolutionized entertainment and recreational activities by introducing innovative applications such as sky painting and writing~\cite{kim2017skywriting,serpiva2021dronepaint}. In addition to this, in the realm of filmmaking and video production, the agility and precision of these UVs enable filmmakers to capture complex scenes in challenging environments. They offer unique perspectives and camera angles, previously unattainable or prohibitively expensive with traditional filming methods, adding a new dimension to storytelling and cinematography~\cite{fleureau2016generic,nageli2018intelligent}. 
    \item \textbf{Heathcare:}
    In healthcare, UAVs have become a game changer. Other than delivering blood and medicine to remote or disaster-affected areas~\cite{thiels2015use,amukele2017drone,claesson2016unmanned}, special design medical drones like Automated External Defibrillator (AED) enabled drones are also deployed to save lives outside hospital~\cite{pulver2016locating,claesson2017drones}. Their ability to quickly and efficiently transport essential medical supplies and equipments has significantly improved access to healthcare services.

    \item \textbf{Industrial Automation :}
Automated vehicles and robots in warehouses significantly reduce manpower requirements and improve logistical processes, demonstrating a smarter approach to internal transport and package sorting~\cite{ogorelysheva2023troubleshooting}. Concurrently, UV swarms are transforming package delivery, providing dynamic and efficient direct-to-customer delivery solutions~\cite{kuru2019analysis,xin2023collaborative,wu2021autonomous,alkouz2020swarm}.

\end{itemize} 

\subsection{Unmanned Aerial Vehicles}

UAVs, also known as drones, constitute a class of aerial robots designed to operate collaboratively to achieve a diverse array of objectives. These objectives span the spectrum from military applications to an expanding suite of civilian and commercial uses. UAVs typically equipped with a multiple of rotors which are capable of VTOL~\cite{TAHIR2019100106}, thus enabling them to hover, ascend, and descend vertically. The operational control of these vehicles may be manual via remote piloting or autonomous, through the integration of sophisticated onboard processing units. Other than the common applications UV swarms offers, the deployment of swarm UAVs in recreational activities has been noteworthy. One interesting application leveraging the flexible movement of UAVs is drone light shows, which involves a fleet of drones, often equipped with lights, flying in a coordinated manner to create shapes, patterns, or texts in the sky~\cite{ang2018high}. Another key application of UAV swarms is in post-disaster communication network reconstruction. Utilizing UAV's mobility, affected areas can quickly reconnect with the outside world by employing UAVs as movable base stations~\cite{akram2020multicriteria,li2019post,merwaday2016improved,malandrino2019multiservice}.

\subsection{Unmanned Ground Vehicles}
UGVs are pivotal components in UV systems, especially for transportation and logistics. Their applications in platoon formation is crucial for enhancing the efficiency and safety of transport systems. Other than classified by the degree of autonomy, UGVs can also be categorized based on their mobility mechanisms into wheeled, tracked, and legged variants~\cite{fernandez2019unmanned}. Tracked UGVs excel in rugged terrains, while legged UGVs are effective in obstacle-rich environments, beneficial for tasks like search and rescue. Besides this, UGVs have been increasingly incorporated into various applications, leveraging their autonomous capabilities for various critical and practical tasks. Other than the aforementioned common applications of UV swarms, collaborations between UAVs and UGVs allow leveraging the strengths of both vehicle types to achieve common goals. For instance, while UAVs can explore areas at a high speed using their cameras, UGVs, capable of traversing a wide variety of terrains, complement UAVs by providing moving recharging stations and transportation between mission objectives, thus enhancing the operational efficiency and extending the mission capabilities~\cite{stolfi2021uav}.

\subsection{Unmanned Surface Vehicles}

USVs also known as autonomous boats, are specialized robotic platforms that operate on water surfaces to perform a variety of tasks. Their operational domain span from disaster management to environmental monitoring and defense applications. Compared to UAV and underwater vehicles, USVs have an advantage in terms of reliable access to Global Positioning System (GPS) data and superior communication capabilities, making them ideal for real-time operations and long-term missions~\cite{liu2020robust}. Such an advantage has make USVs instrumental in conducting water depth surveys and examining mitigation patterns and changes within major ecosystems~\cite{HEINS2017749}. USVs also facilitate research activities that require heterogeneous UV cooperation, including integration with other UVs~\cite{s23104643}. With the ability to generate power from solar, wind, and waves, USVs can serve as durable platforms for collecting extensive data across various locations and act as mobile communication and refueling relays for other vehicles~\cite{s19030702,LIU201671}.

\subsection{Unmanned Underwater Vehicles}

UUVs, also known as submersible drones, are specialized aquatic robots designed for a wide range of underwater tasks. Initially focused on military and scientific uses, their applications have expanded into civilian and recreational areas~\cite{s23177321, neira2021review}. UUVs are categorized into Remotely Operated Vehicles (ROVs), connected to operators via cables and used for detailed tasks like observation and maintenance, and Autonomous Underwater Vehicles (AUVs), which operate independently for various measurements, including submarine volcano monitoring and seabed surveys~\cite{liu2022remotely, neira2021review}. A notable application of UUVs is in inspecting and maintaining marine structures, providing essential data for the construction industry and ensuring the integrity of underwater infrastructure, such as oil pipelines and undersea cables~\cite{s20216203}. Additionally, UUVs are uniquely suited for extreme environments, such as ice-covered regions, nuclear facilities, and other areas where human access is dangerous or impossible, showcasing their versatility and critical role in challenging underwater operations~\cite{s20216203,s23177321}.

UVs are revolutionizing various industries by simplifying traditional operations and innovating new activities in air, land, sea, and underwater domains. The increasing number of UVs sets a foundation for generative AI to enhance, promising to significantly improve autonomy and coordination among swarms.
\section{Generative AI}
\label{GAI}
% https://arxiv.org/pdf/2303.04226.pdf

\subsection{An Overview}
\begin{figure*}[t!]
\centerline{\includegraphics[width=0.9\textwidth]{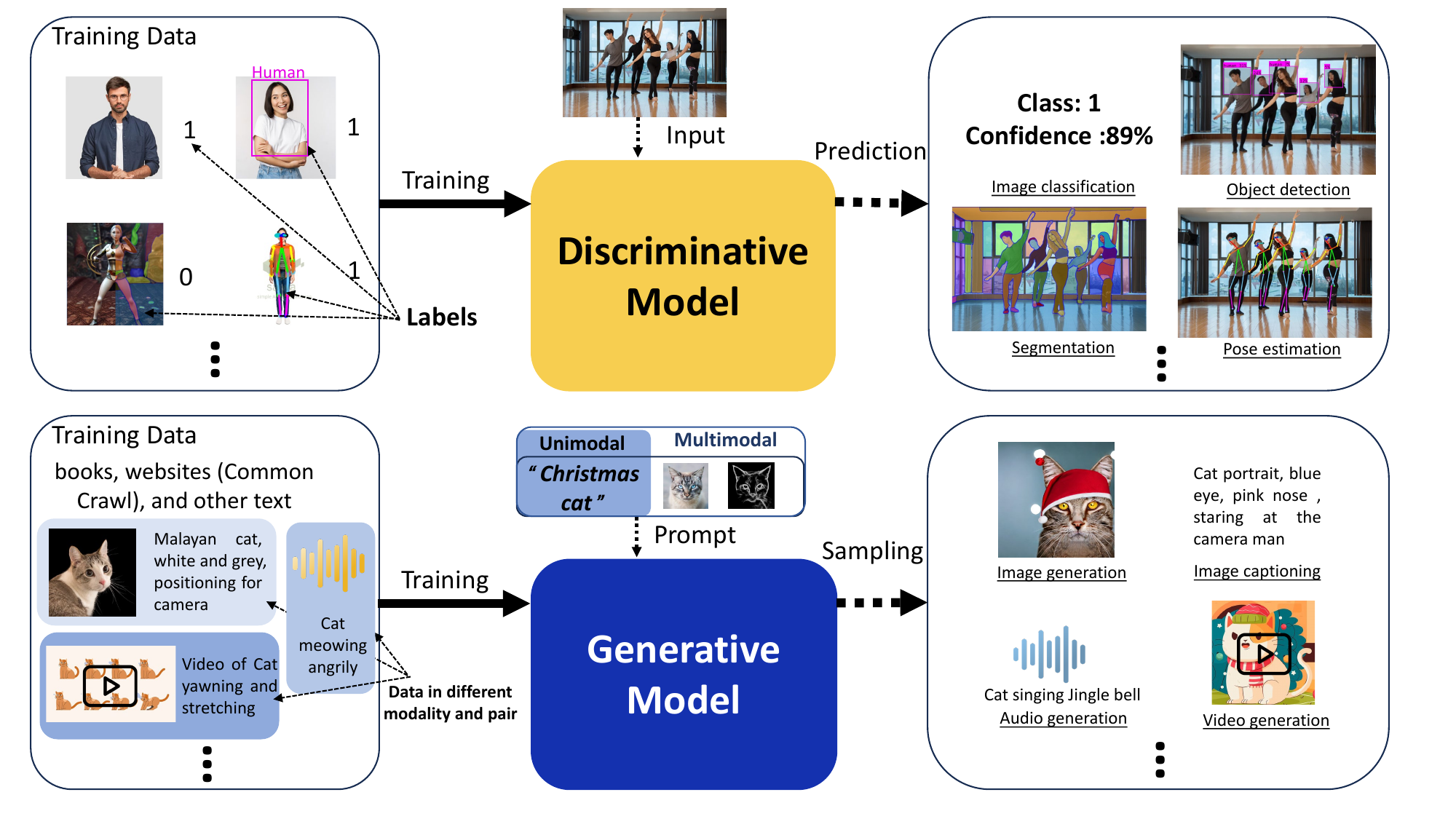}}
\caption{Generative AI vs Discriminative AI}
\label{GENVSDIS}
\end{figure*}

GAI represents a paradigm shift in AI technology, characterized by its ability to produce novel and meaningful content, such as text, images, audio, and 3D models. As illustrated in Fig.~\ref{GENVSDIS}, unlike discriminative models that focus on classification or prediction, GAI models are adept at interpreting instructions and generating tangible outputs, a distinction that marks a significant leap in AI capabilities~\cite{byrne2023disruptive}. These advanced models could capture a deep understanding of data patterns and structures, enabling them to not only replicate but also innovate within the learned frameworks. This innovation is evident in GAI's diverse applications, ranging from realistic image~\cite{zhang2023text} and text creation~\cite{iqbal2022survey} to complex 3D model generation~\cite{regenwetter2022deep}.

GAI is revolutionizing traditional methods, offering transformative impacts in domains such as medical and engineering education through personalized learning support and intelligent tutoring systems~\cite{bahroun2023transforming}. Similarly, in visual content generation, GAI's potential for causal reasoning is being explored, the ability to reason about causality is crucial for many applications, such as robotics, autonomous driving, and medical diagnosis~\cite{goyal2017making,li2023image}.

Moreover, GAI's influence extends to business model innovation, where its applications in various industries, including software engineering, healthcare and financial services, are reshaping traditional business models~\cite{kanbach2023genai}. This versatility of GAI not only underscores its role as a tool for creativity and innovation, but also highlights its potential to drive significant advancements across multiple sectors. In summary, the primary distinction between GAI and traditional Discriminative lies in their capabilities, i.e., GAI is designed to create and innovate, generating new content, while traditional Discriminative is more focused on data classification and extraction~\cite{du2023spear}.

\subsection{Typical GAI Models}

\begin{figure*}[t!]
\centerline{\includegraphics[width=0.9\textwidth]{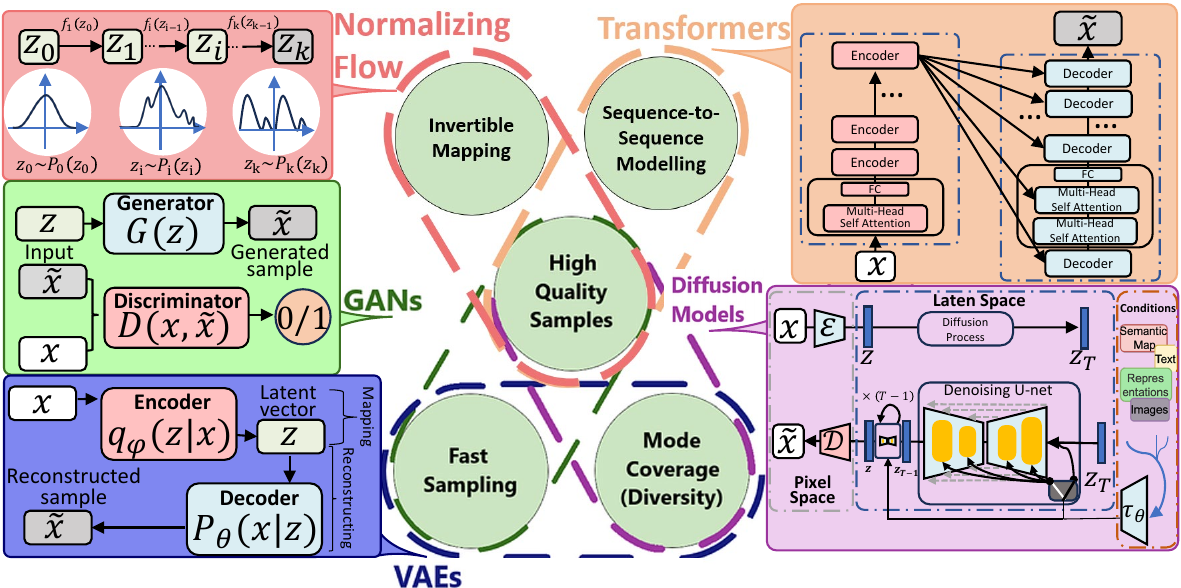}}
\caption{Overviews of Generative Models: Common Abilities and Structural Overviews of GANs, VAEs, GDMs, Normalizing Flows, and Transformers}
\label{GENVdiverse}
\end{figure*}

\subsubsection{GAN}
GANs represent a significant advancement in both semi-supervised and unsupervised learning. Conceptualized by Goodfellow~\cite{goodfellow2014generative} in 2014, GANs involve training two networks simultaneously: a generator and a discriminator. The generator's role is to produce data that mimics real data, while the discriminator acts as a classifier, distinguishing between real and generated data. This dynamic between the two networks forms the core of the GAN model, where the generator aims to create data realistic enough to confuse the discriminator, and the discriminator improves its ability to identify fake data. This process results in a Nash equilibrium, where the generator produces increasingly realistic data, and the discriminator becomes more adept at identifying forgeries. This effectively utilizes a supervised learning approach to achieve unsupervised learning outcomes by generating synthetic data that appears authentic~\cite{goodfellow2020generative,jabbar2021survey}.
Despite these achievements, GAN training remains challenging, primarily due to the instability. This is critical as the generator and discriminator in GANs need to optimize through alternating or simultaneous gradient descent. The GAN architecture faces several issues like achieving Nash equilibrium~\cite{ratliff2013characterization}, mode collapse and vanishing gradients~\cite{goodfellow2016nips}. To address these challenges, numerous solutions have been proposed, such as unrolled GAN~\cite{metz2016unrolled}, mini-batch discrimination, historical averaging, feature matching~\cite{salimans2016improved}, two timescale update rule~\cite{heusel2017gans} and self-attention GAN~\cite{zhang2019self}. These developments have been crucial in stabilizing GAN training over the years.

As shown in Fig. \ref{GENVdiverse}, GAN excel in generating high-quality samples and achieving fast sampling primarily due to their unique adversarial training mechanism. In a GAN, the generator and discriminator networks engage in a continuous competitive process, where the generator learns to produce increasingly realistic samples to deceive the discriminator. In UV swarm applications, the competitive process between the generator and discriminator in GANs not only ensures the generation of realistic samples but also aids in the creation of varied and complex environmental simulations crucial for training UVs. Moreover, the efficiency of GANs in sample generation becomes significant when considering the computational constraints and the need for rapid decision-making in UV swarms. A trained generator in GAN is able to produce new samples through simple forward inferences. Lastly, GANs' capacity to learn a rich and diverse latent space is essential for UV swarms. This ability allows for the generation of varied scenarios and conditions, which is crucial for the robust training of these systems. In summary, the unique features of GANs such as high-quality sample generation, efficiency in producing new samples, and the ability to learn a rich, diverse latent space are particularly beneficial for UV swarms, enhancing their adaptability, efficiency, and reliability in dynamic and challenging environments.

\subsubsection{VAE}
VAEs are deep latent space generative models that fundamentally learn the distribution of data to generate new, meaningful data with more intra-class variations. Similar to GANs, VAEs consist of two interlinked yet independently parameterized components: the encoder and the decoder. The encoder provides the decoder with an estimation of its posterior over latent variables. This estimation is crucial for the decoder to update its parameters during iterations of ``expectation maximization” learning. Conversely, the decoder forms a framework that assists the encoder in learning meaningful data representations, which may include class-labels. The encoder essentially serves as an approximate inverse to the generative model, in line with Bayes' rule~\cite{kingma2019introduction}. The training of VAEs involves optimizing the Evidence Lower Bound (ELBO), which balances reconstruction accuracy and the similarity of the latent space distribution with the target distribution~\cite{yang2017understanding}.

In the context of data augmentation, VAEs are valuable for their ability to increase the variance of a dataset, particularly in domains with limited ideal training samples~\cite{kingma2019introduction}. Although VAE can avoid issues like non-convergence and mode collapse, which are common in other generative models like GANs, the samples generated by VAEs tend to be of lower quality compared to GANs. Representation learning is another significant application of VAEs. This approach involves transforming raw data into more advanced training data representations, often requiring significant human expertise and effort. VAEs automate this process by learning mappings from high-dimensional space to a meaningful low-dimensional embedding~\cite{leelarathna2023enhancing}.

In UV swarm applications, VAEs stand out for their stability and reliability. Compared to GANs, VAEs can mitigate issues like mode collapse, making them a more stable choice for generating training simulations~\cite{srivastava2017veegan}. This stability is crucial in UV swarms, where consistent and varied environmental modeling is essential for thorough system training. The robustness of VAEs in generating a wide range of scenarios without the risk of model collapse ensures that UV systems are exposed to a comprehensive set of conditions, enhancing their adaptability and preparedness for real-world operations.
\subsubsection{GDM}
Different from GANs and VAEs, GDM utilize a two-stage process involving both forward and reverse diffusion~\cite{cao2022survey,yang2023diffusion}. A diffusion model is a parameterized Markov chain trained using variational inference to produce samples matching the data after finite time~\cite{ho2020denoising}. In the forward diffusion stage, these models gradually add gaussian noise to input data over multiple steps, progressively degrading the data's structure. During the reverse diffusion stage, the model learns to methodically reverse this process, sequentially predicting under guidance of prompt and removing the noise and thereby reconstructing a new data sample. The removed noise at each step is estimated via a neural network like U-Net architecture to ensure dimension preservation~\cite{li2023faster}.

Research on diffusion models has shown promising results in various computer vision tasks, with three primary subcategory being Denoised Diffusion Probabilistic Models (DDPM), Noise-Conditioned Score Network (NCSN), and Stochastic Differential Equations (SDE)~\cite{croitoru2023diffusion}. DDPMs excel in creating diverse and high-quality images and serve as the foundational structure for well-known models like Stable Diffusion ~\cite{rombach2022high} and DALL-E series~\cite{ramesh2021zero,ramesh2022hierarchical,betker2023improving}. NCSNs are the core technology of Deepfake~\cite{westerlund2019emergence}, which is deployed to produce realistic altered images and videos through score matching and noise level training. SDEs utilize forward and reverse SDEs for robust and theoretically sound generation strategies are often applied to models like DiffFlow~\cite{zhang2023diffflow}. These models have achieved remarkable results in image generation, surpassing GANs in diversity of generated samples, but the need for multiple adding noise and denoising steps during inference makes them slower than GANs and less efficient than VAEs in image production.  To address the efficiency challenge in GDM, a key research focus is on enhancing sampling efficiency. An example is the development of Denoising Diffusion Implicit Models (DDIM)~\cite{song2020denoising}. DDIMs improve the sampling speed by allowing fewer steps in generating samples without significantly compromising the quality of the generated images. 

In the context of UV swarms, GDMs can be particularly useful in generating highly detailed and diverse environmental simulations for training UV systems~\cite{katara2023gen2sim}. Unlike GANs that often suffer from mode collapse, or VAEs that sometimes generate low quality images, GDMs can produce samples with a high level of detail and variation. This is crucial for UV swarms to train in realistic and varied conditions~\cite{zhang2023generative}. On the other hand, the iterative process of GDMs that involves multiple steps of noise addition and removal, results in a slower and less efficient sample generation compared to GANs and VAEs. However, this trade-off is justifiable in scenarios such as search and rescue operations, military reconnaissance, and environmental monitoring, where the utmost realism and detail in training simulations are paramount for the effective functioning of UV swarms.

\subsubsection{Transformer}
Transformers models~\cite{vaswani2017attention} have become a fundamental in numerous state-of-the-art models in the field of generative models. Especially in natural language processing, Transformer-based Large Language Models (LLMs) such as GPT series~\cite{brown2020language,radford2018improving,radford2019language}, Bidirectional Encoder Representations from Transformers (BERT)~\cite{devlin2018bert} and Bard\footnote{https://ai.google/static/documents/google-about-bard.pdf , accessed Jan 3rd,2024} have demonstrated their abilities to capture large corpora of information~\cite{karapantelakis2023generative}. Although both VAE and Transformer architectures feature an encoder-decoder design, their functionalities diverge significantly. In transformers, the encoder processes the input sequence, capturing complex dependencies, and the decoder generates the output sequence, often leveraging self-attention mechanisms to focus on different parts of the input data. Both the encoder and decoder in transformers consist of multiple layers of attention mechanisms and feed-forward neural networks, enabling them to handle complex sequence data effectively~\cite{niu2021review}.

Transformers can offer more than just standalone models. When integrated into other generative models, they introduce essential mechanisms and techniques such as attention, self-attention, multi-head attention, and positional encoding~\cite{khan2022transformers}. This integration has unlocked practical applications of transformers in various domains, including text generation for creative writing~\cite{marco2022systematic}, chatbots~\cite{masum2021transformer}, code generation~\cite{svyatkovskiy2020intellicode}, and programming assistance~\cite{gao2023code}. Notably, incorporating transformers into other GAI has led to significant advancements in image synthesis. Pure transformer-based architectures in GANs, such as ViTGAN~\cite{lee2021vitgan} and STrans GAN~\cite{xu2021stransgan}, have successfully synthesized high-resolution images without the need for convolutions neural networks. These exciting developments demonstrate the versatility and expanding capabilities of transformers in generative tasks, extending their utility from text-based applications to sophisticated image synthesis.
    \begin{figure*}[t!]
\centerline{\includegraphics[width=0.9\textwidth]{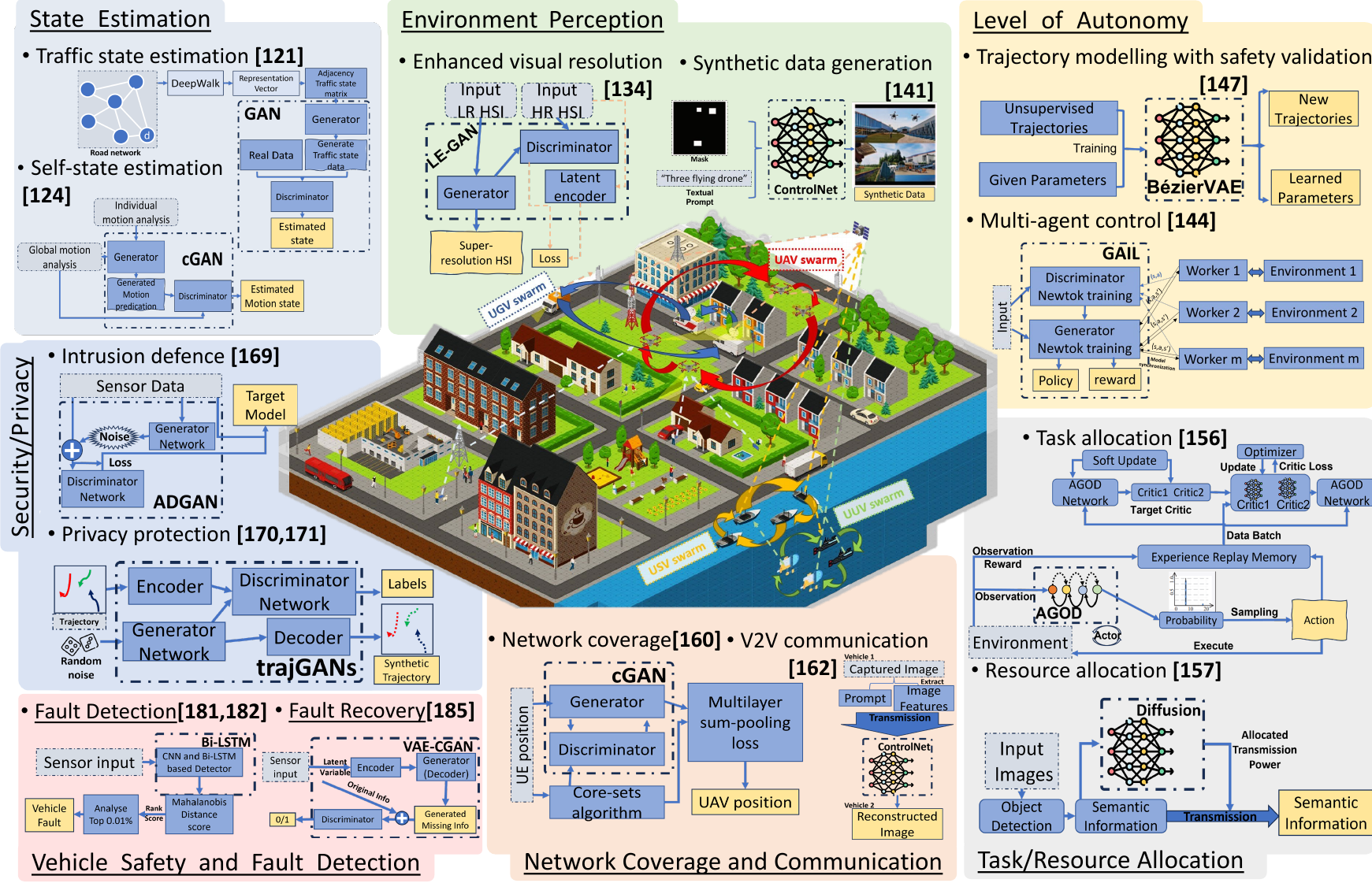}}
\caption{Exploring the Spectrum of Innovation: This illustration presents 12 groundbreaking model structures, featuring two distinct approaches per aspect, to demonstrate the diverse applications of GAI in enhancing the performance and addressing challenges in UV swarms. Each model encapsulates unique strategies and solutions, offering a comprehensive overview of the technological advancements in this field.}
\label{systemdigram}
\end{figure*}
In the realm of UV swarms, the transformer architecture has become a key asset due to its proficiency in processing sequential data and managing long-range dependencies, greatly enhancing tasks that require intricate decision-making based on comprehensive data streams~\cite{singh2023time}. In UV swarm operations, each unit may need to process and respond to a vast array of sensor inputs, communications from other units, and environmental factors, making the transformer's ability to analyze this sequence data and identify crucial dependencies a critical resource in aiding the rapid decision-making process~\cite{volger2015human}. Furthermore, the generative capabilities of transformers have shown great potential to create highly detailed and context-aware simulations for training or mission planning in UV swarms. Models such as GPT-4 could prove instrumental in generating realistic, complex scenarios for training UVs and enhancing their preparedness for real-world operations with their enormous parameter count and the ability to handle both text and image prompts~\cite{liao2023gpt}. 

\subsubsection{Normalizing flow}
Normalizing Flows (NF) are a class of generative models that stand out due to their capability to produce tractable distributions where both sampling and density evaluation can be efficient and exact~\cite{kobyzev2020normalizing}. They are characterized by transforming a simple probability distribution, such as a standard normal distribution, into a more complex one through a sequence of invertible and differentiable mappings. This transformation allows for the evaluation of the density of a sample by reverting it back to the original simple distribution and calculating the product of the density of the inverse-transformed sample and the associated change in volume induced by the sequence of inverse transformations~\cite{coeurdoux2023normalizing}.

A key advantage of NFs over other generative models is their inherent invertibility. This feature enables exact reverse mapping, which is crucial for efficiently and accurately evaluating the density of generated samples – a challenging task in other generative models like GANs or VAEs~\cite{reyes2023testing}. The invertibility of NFs also enables both efficient sampling and exact density estimation, making them highly versatile. Unlike GANs, which may suffer from training instabilities such as mode collapse, NFs offer a more stable training process~\cite{liu2020empirical}. Additionally, NFs can reconstruct data with higher fidelity compared to the approximate reconstruction in VAEs~\cite{kobyzev2020normalizing}.

NFs have been deployed in various applications, including image generation~\cite{ho2019flow++}, noise modelling~\cite{abdelhamed2019noise}, video generation~\cite{kumar2019videoflow}, audio generation~\cite{esling2019universal}, and graph generation~\cite{madhawa2019graphnvp}. In the context of UV swarm, Graphical Normalizing Flows (GNF) models can be utilized for anomaly detection in UV swarms~\cite{ma2023detecting}. GNF leverages Bayesian networks to identify relationships among time series components and perform density estimation in the low-density regions of a distribution where anomalies typically occur. Comparing to other GAI models, the key advantage of NF models lies in their capability to provide exact likelihood estimation, which offers a precise, stable, and efficient solution for identifying and responding to various anomalies in UV systems~\cite{zhang2023generative}.
\section{Solutions to Unmanned Vehicle swarms utilizing GAI}
\label{survey}

\subsection{State Estimation}
\begin{table*}[!ht]
\renewcommand{\multirowsetup}{\centering}

\caption{Generative AI in State Estimations}
\vspace*{2mm}
\footnotesize
\centering
\renewcommand\arraystretch{1.27}
\begin{tabular}{c||c<{\centering}||m{0.375\textwidth}|m{0.175\textwidth}|m{0.175\textwidth}}
\hline
\hline
\textbf{Type}&\textbf{Ref}&\textbf{Description}&\textbf{Pros}&\textbf{Cons} \\
\hline
\hline
\multirow{4}{*}{\textbf{GAN}}&\cite{XU2020102635}&
GAN is deployed to estimate traffic states by generating realistic and diverse samples from sparse and incomplete traffic data.&
\begin{itemize} [leftmargin=3pt,rightmargin=0pt,after=\vspace{-1\baselineskip}]
    \item Accurate spatial-temporal correlation capture and
    \item Adapts to dynamic changes
\end{itemize}&
\begin{itemize} [leftmargin=3pt,rightmargin=0pt,after=\vspace{-1\baselineskip}]
    \item High memory and computing needs
    \item Limited generalization
\end{itemize} \\
\cline{2-5}
&\cite{10135862}&
GAN is used in this paper to construct a harmonic state estimation model, where the mechanism equation is integrated into the objective function to include power grid information.&
\begin{itemize}[leftmargin=3pt,rightmargin=0pt,after=\vspace{-1\baselineskip}]
    \item Accurate source localization
\end{itemize}&
\begin{itemize}[leftmargin=3pt,rightmargin=0pt,after=\vspace{-1\baselineskip}]
    \item Potential GAN-related issues like model collapse
\end{itemize} \\
\cline{2-5}
&\cite{8914585}&
Conditional GAN (cGAN) is employed to refine state estimates by learning the mapping between raw estimates and true states.&
\begin{itemize}[leftmargin=3pt,rightmargin=0pt,after=\vspace{-1\baselineskip}]
    \item Improved estimation accuracy
\end{itemize}&
\begin{itemize}[leftmargin=3pt,rightmargin=0pt,after=\vspace{-1\baselineskip}]
    \item Sensitive to hyperparameters change
    \item Large training data required
\end{itemize} \\
\cline{2-5}
&\cite{yu2020conditional}&
cGANs are employed to fuse individual and global motion for enhanced multiple object tracking.&
\begin{itemize}[leftmargin=3pt,rightmargin=0pt,after=\vspace{-1\baselineskip}]
    \item Enhanced tracking accuracy
\end{itemize}&
\begin{itemize}[leftmargin=3pt,rightmargin=0pt,after=\vspace{-1\baselineskip}]
    \item Computational overhead
    \item Training data dependency
\end{itemize} \\
\hline
\multirow{1}{*}{\textbf{VAE}}&\cite{10189383}&
VAE is utilized to capture the temporal correlations in the wireless channels of UAVs for improved channel state estimation.&
\begin{itemize}[leftmargin=3pt,rightmargin=0pt,after=\vspace{-1\baselineskip}]
    \item Reduce signal transmission needs
\end{itemize}&
\begin{itemize}[leftmargin=3pt,rightmargin=0pt,after=\vspace{-1\baselineskip}]
    \item Increased computational resources
    \item Model hyperparameter sensitivity
\end{itemize} \\
\hline
\multirow{1}{*}{\textbf{Diffusion}}&\cite{9957135}&
Diffusion-based score models are employed to exploit the natural correlations in MIMO channels for effective state estimation.&
\begin{itemize}[leftmargin=3pt,rightmargin=0pt,after=\vspace{-1\baselineskip}]
    \item Enhanced channel estimation
\end{itemize}&
\begin{itemize}[leftmargin=3pt,rightmargin=0pt,after=\vspace{-1\baselineskip}]
    \item Computationally intensive
    \item Noise level impact on accuracy
\end{itemize} \\
\hline
\multirow{1}{*}{\textbf{Others}}&\cite{delecki2023deep}&
Deep normalizing flows are employed to capture the structure and intricacies of the state distribution for state estimation.&
\begin{itemize}[leftmargin=3pt,rightmargin=0pt,after=\vspace{-1\baselineskip}]
    \item Flexible complex distribution modeling
\end{itemize}&
\begin{itemize}[leftmargin=3pt,rightmargin=0pt,after=\vspace{-1\baselineskip}]
    \item High computational demand
    \item Large data requirement
\end{itemize} \\
\hline
\end{tabular}
\label{Statetable}
\end{table*}
State estimation is critical to UVs swarm applications, especially in fields like autonomous driving and traffic estimation. State variables like position, velocity, and orientation, play a crucial role in lateral's decision making during navigation or trajectory planning~\cite{barfoot2017state}. However, the stochastic nature of system measurements and robot dynamics can lead to uncertainty about the actual state. Therefore, the primary objective of state estimation is to deduce the distribution over state variables based on the available time observations~\cite{delecki2023deep}.

As shown in Table~\ref{Statetable}, the integration of GAI in state estimation for UVs offers a broad range of innovative methodologies, each tailored to specific challenges and operational contexts. For instance, in addressing the challenge of data insufficiency in traffic state estimation for UGVs, the authors in~\cite{XU2020102635} utilized Graph-Embedding GANs to generate realistic traffic data for underrepresented road segments by capturing the spatial interconnections within road networks. In this proposed framework, the generator uses embedded vectors from similar road segments to simulate real traffic data. Meanwhile, the discriminator distinguishes between this synthesized and actual data and iteratively train generator to optimize both components until the generated data is statistically indistinguishable from real data. This methodology not only fills data gaps but also significantly enhances estimation accuracy, as evidenced reduction in mean absolute error compared to traditional models like Deeptrend2.0~\cite{DAI2019142}. Such an advancement in traffic state estimation underscores GAI's potential in improving UGV navigation and decision-making in complex traffic scenarios~\cite{XU2020102635}.

In addition to the standard GAN, cGANs can be used to generate the corresponding estimated system state variable given the raw measurements~\cite{8914585}. The challenge of accurately estimating the motion of multiple UAVs in dynamic environments is addressed using a cGAN framework by employ raw measurements of sensor as conditional constrain. The authors in~\cite{yu2020conditional} combine individual motion predictions from a Social LSTM network~\cite{alahi2016social} with global motion insights from a Siamese network~\cite{he2018twofold} to achieve a comprehensive motion state prediction. This method excels in accurately forecasting UAV trajectories which is crucial for effective swarm navigation. By effectively disentangling and fusing individual and global motions, the cGAN based framework demonstrates performing well and improving the performance of multiple object tracking compared to the original Social LSTM.
\begin{table*}[!ht]
\renewcommand{\multirowsetup}{\centering}
\caption{Generative AI in Environmental Perception}
\footnotesize
\centering
\renewcommand\arraystretch{1.27}
\begin{tabular}{c||c<{\centering}||m{0.375\textwidth}|m{0.175\textwidth}|m{0.175\textwidth}}
\hline
\hline
\textbf{Type}&\textbf{Ref}&\textbf{Description}&\textbf{Pros}&\textbf{Cons} \\
\hline
\hline
\multirow{4}{*}{\textbf{GAN}} &~\cite{shi2022latent} &
Latent Encoder Coupled GAN (LE-GAN) is introduced to achieve efficient hyper-spectral image super-resolution.
&
\begin{itemize}[leftmargin=3pt,rightmargin=0pt,after=\vspace{-1\baselineskip}]
\item Effective for image super-resolution.
\end{itemize}
&
\begin{itemize}[leftmargin=3pt,rightmargin=0pt,after=\vspace{-1\baselineskip}]
\item Risk of mode collapse and spectral-spatial distortions.
\end{itemize}
\\
\cline{2-5}
&~\cite{8569971} &
Trajectory GAN (TraGAN) is used to create realistic and intuitive lane change trajectories from recorded highway traffic data.
&
\begin{itemize}[leftmargin=3pt,rightmargin=0pt,after=\vspace{-1\baselineskip}]
\item Learns trajectory parameters automatically without labels.
\end{itemize}
&
\begin{itemize}[leftmargin=3pt,rightmargin=0pt,after=\vspace{-1\baselineskip}]
\item Needs large, high-quality datasets and sensitive to hyperparameters.
\end{itemize}
\\
\cline{2-5}
&~\cite{zhang2018deeproad} &
The DeepRoad framework, which deploys GANs for metamorphic testing and input validation in autonomous driving systems is proposed.
&
\begin{itemize}[leftmargin=3pt,rightmargin=0pt,after=\vspace{-1\baselineskip}]
\item Improves the reliability of testing by addressing scene diversity.
\end{itemize}
&
\begin{itemize}[leftmargin=3pt,rightmargin=0pt,after=\vspace{-1\baselineskip}]
\item Synthetic images might be insufficient due to limitations.
\end{itemize}
\\
\cline{2-5}
&~\cite{zharkovsky2020end} &
A cGAN-based framework is deployed for automatic change detection in UAV and remote sensing images.
&
\begin{itemize}[leftmargin=3pt,rightmargin=0pt,after=\vspace{-1\baselineskip}]
\item Shows promising results in automatic change detection.
\end{itemize}
&
\begin{itemize}[leftmargin=3pt,rightmargin=0pt,after=\vspace{-1\baselineskip}]
\item Potential need for large training data sets.
\end{itemize}
\\
\hline
\multirow{4}{*}{\textbf{VAE}} &~\cite{8569971} &
Trajectory VAE (TraVAE) is used to generate synthetic lane change trajectories data from real traffic record.
&
\begin{itemize}[leftmargin=3pt,rightmargin=0pt,after=\vspace{-1\baselineskip}]
\item Learns intuitive latent parameters for lane changes.
\end{itemize}
&
\begin{itemize}[leftmargin=3pt,rightmargin=0pt,after=\vspace{-1\baselineskip}]
\item Higher reconstruction error than TraGAN.
\end{itemize}
\\
\cline{2-5}
&~\cite{ISLAM2021105950} &
VAE is deployed to generate synthetic crash data from real traffic data to augment the dataset.
&
\begin{itemize}[leftmargin=3pt,rightmargin=0pt,after=\vspace{-1\baselineskip}]
\item Produces realistic and diverse crash data.
\end{itemize}
&
\begin{itemize}[leftmargin=3pt,rightmargin=0pt,after=\vspace{-1\baselineskip}]
\item May require strong assumptions leading to suboptimal models.
\end{itemize}
\\
\cline{2-5}
&~\cite{9785931} &
Image translation framework based on VAE and GANs are used to translate simulated images into realistic synthetic images that can be used for training and testing change detection models.
&
\begin{itemize}[leftmargin=3pt,rightmargin=0pt,after=\vspace{-1\baselineskip}]
\item Generates labeled data with various imaging challenges.
\end{itemize}
&
\begin{itemize}[leftmargin=3pt,rightmargin=0pt,after=\vspace{-1\baselineskip}]
\item Real images still needed as references
\item May not cover all domain gaps.
\end{itemize}
\\
\cline{2-5}
&~\cite{10134962} &
The VAE encode the environment image from pixel space to a smaller dimensional latent space before the diffusion progress and after the diffusion progress the VAE generates the final image by converting the representation into pixel space.
&
\begin{itemize}[leftmargin=3pt,rightmargin=0pt,after=\vspace{-1\baselineskip}]
\item Compressing high-dimensional data into a lower-dimensional latent space.
\item Reduce the computational resources required.
\end{itemize}
&
\begin{itemize}[leftmargin=3pt,rightmargin=0pt,after=\vspace{-1\baselineskip}]
\item Struggle to accurately reconstruct the finer details.
\end{itemize}
\\
\hline
\multirow{2}{*}{\textbf{Diffusion}} &~\cite{10134962} &
Stable diffusion that is developed base on GDM is deployed to generate synthetic images of the UAV models.
&
\begin{itemize}[leftmargin=3pt,rightmargin=0pt,after=\vspace{-1\baselineskip}]
\item Captures high-level UAV model features and fine details.
\end{itemize}
&
\begin{itemize}[leftmargin=3pt,rightmargin=0pt,after=\vspace{-1\baselineskip}]
\item Computationally expensive and slow to train.
\end{itemize}
\\
\cline{2-5}
&~\cite{Synthetic2023} &
Conditional GDM are deployed to generates photo-realistic images and corresponding ground truth bounding boxes for UAV detection according to conditional inputs, such as binary masks that specify the details and background of the UAVs, and text prompts that describe the scenes.
&
\begin{itemize}[leftmargin=3pt,rightmargin=0pt,after=\vspace{-1\baselineskip}]
\item Boosts UAV detector performance with high-fidelity images.
\end{itemize}
&
\begin{itemize}[leftmargin=3pt,rightmargin=0pt,after=\vspace{-1\baselineskip}]
\item Potential interference in object detection if not well-filtered.
\end{itemize}
\\
\hline
\multirow{2}{*}{\textbf{Others}} &~\cite{drones7020114} &
To generate captions for images captured by UAVs using CLIP Prefix for Image Captioning, a transformer-based architecture that uses the CLIP and GPT-2 models is used.
&
\begin{itemize}[leftmargin=3pt,rightmargin=0pt,after=\vspace{-1\baselineskip}]
\item Translates visual content to coherent text descriptions.
\end{itemize}
&
\begin{itemize}[leftmargin=3pt,rightmargin=0pt,after=\vspace{-1\baselineskip}]
\item May produce erroneous or inconsistent text descriptions.
\end{itemize}
\\
\cline{2-5}
&~\cite{rs15092221} &
Generative knowledge-supported transformer (GKST) that leverages the mutual learning across different views is deployed to improve the feature representation ability and retrieval performance.
&
\begin{itemize}[leftmargin=3pt,rightmargin=0pt,after=\vspace{-1\baselineskip}]
\item Bridges appearance gap between ground and aerial views.
\end{itemize}
&
\begin{itemize}[leftmargin=3pt,rightmargin=0pt,after=\vspace{-1\baselineskip}]
\item Computationally expensive.
\item Memory-intensive.
\end{itemize}
\\
\hline
\end{tabular}
\label{Envirtable}
\end{table*}
Additionally, an application of VAEs in capturing temporal correlations in UAV wireless channels underscores the importance of GAI in communication systems, improving channel state estimation and signal clarity by generating realistic and diverse channel samples~\cite{10189383}. The exploration extends to diffusion-based score models and deep normalizing flows, employed for generating complex state variable distributions, showcasing the capability of GAI to model and estimate states in more flexible manners ranging from state variables (i.e., position, velocity, and orientation) to the intricate high-dimensional gradient of these distributions~\cite{9957135,delecki2023deep}.

The versatility of GAI in state estimation for UV swarms is evident in two aspects: ability of generating missing information through adversarial mechanisms and ability of fusing varied data sources for comprehensive state analysis. These ability enables more accurate states estimation in complex operational scenarios.

\subsection{Environmental Perception} 

Environmental perception in the context of UVs typically refers to the ability of the vehicle to perceive and understand its surrounding environment in real-time~\cite{zhang2021research}. This is a key technology for achieving autonomous navigation and completing tasks for UV swarms . Such technology often involves the use of sensors such as LiDAR, cameras, and millimeter wave radar to interact with the external environment~\cite{zhang2023research}. The realm of environmental perception in UVs is markedly advanced by the varied and innovative applications of GAI, as detailed in Table \ref{Envirtable}. For example, due to intrinsic constraints, such as motion blur from movement, adverse weather conditions, and varying flying altitudes, UAVs often capture low-resolution images. To address this problem, the authors in~\cite{shi2022latent} introduce a framework called Latent Encoder Coupled Generative Adversarial Network (LE-GAN), designed for efficient hyper-spectral image (HSI) super-resolution. The generator in LE-GAN uses a short-term spectral-spatial relationship window mechanism to exploit the local-global features and enhance the informative band features. The discriminator adopts Wasserstein distance-based loss between the probability distributions of real and generated images. Such a framework not only improves the SR quality and robustness, but also alleviates the spectral-spatial distortions caused by mode collapse problem by learning the feature distributions of high resolution HSIs in the latent space~\cite{shi2022latent}. 

 Besides improving UVs' accuracy by enhancing remote sensing resolution, a more common application of GAI is to generate synthetic datasets, which indicates the challenge of reduced model accuracy caused by insufficient data~\cite{10134962}. For instance, a framework named Trajectory GAN (TraGAN) is deployed for generating realistic lane change trajectories from highway traffic data~\cite{8569971}. Another GAN-based framework named DeepRoad is utilized for testing and input validation in autonomous driving systems~\cite{zhang2018deeproad} to enhance the reliability of testing by generating driving scenes in different weather conditions. VAEs are also deployed for generating more realistic and diverse crash data, addressing the limitations of traditional data augmentation methods~\cite{ISLAM2021105950}. Additionally, image translation frameworks that combine VAE and GANs are utilized for transforming simulated images into realistic synthetic ones for training and testing change detection models~\cite{zharkovsky2020end,9785931}, though they still require real images for reference. Moreover, the authors in~\cite{Synthetic2023} introduce a method leveraging a text-to-image diffusion model to generate realistic and diverse images of UAVs set against various backgrounds and poses. With more than 20,000 synthetic images generated by merging background descriptions and binary masks based on ground truth bounding boxes, the detector's average precision on real-world data increased 12\%.
\begin{table*}[!ht]
\renewcommand{\multirowsetup}{\centering}
\caption{Generative AI in Level of Autonomy}
\footnotesize
\centering
\renewcommand\arraystretch{1.27}
\begin{tabular}{c||c<{\centering}||m{0.375\textwidth}|m{0.175\textwidth}|m{0.175\textwidth}}
\hline
\hline
\textbf{Type}&\textbf{Ref}&\textbf{Description}&\textbf{Pros}&\textbf{Cons} \\
\hline
\hline
\multirow{3}{*}{\textbf{GAN}} &~\cite{xiang2022flock, song2018multi} & 
 Generative Adversarial Imitation Learning (GAIL) is integrated with multi-agent DRL to enhance cooperative search strategies for UAVs which allows UAVs to learn efficient searching strategies by imitating expert behaviors. &
\begin{itemize}[leftmargin=3pt,rightmargin=0pt,after=\vspace{-1\baselineskip}]
    \item Simplifies learning process without explicit rewards.
    \item Potentially leads to more natural and efficient behaviors.
\end{itemize} &
\begin{itemize}[leftmargin=3pt,rightmargin=0pt,after=\vspace{-1\baselineskip}]
    \item Requires significant expert trajectory data for training.
\end{itemize} \\
\cline{2-5}
&~\cite{dronegial2023} & 
GAIL is utilized to train UAVs for navigation tasks within the virtual environment. The UAVs learn to navigate by imitating expert trajectories, enabling them to understand and adapt to complex and dynamic scenes within the virtual environment.  &
\begin{itemize}[leftmargin=3pt,rightmargin=0pt,after=\vspace{-1\baselineskip}]
    \item Improves generalization in various virtual scenarios.
\end{itemize} &
\begin{itemize}[leftmargin=3pt,rightmargin=0pt,after=\vspace{-1\baselineskip}]
    \item Time-consuming collection of expert demonstrations.
\end{itemize} \\
\hline
\multirow{3}{*}{\textbf{VAE}} &~\cite{BézierVAE2019safetytrj} & 
BézierVAE is utilized for modeling vehicle trajectories, especially for safety validation in highly automated driving scenarios. The method encodes trajectories into a latent space using VAEs and then decodes them using Bézier curves to reconstruct and generate new trajectories.  &
\begin{itemize}[leftmargin=3pt,rightmargin=0pt,after=\vspace{-1\baselineskip}]
    \item Captures and generates diverse driving behaviors.
\end{itemize} &
\begin{itemize}[leftmargin=3pt,rightmargin=0pt,after=\vspace{-1\baselineskip}]
    \item Requires diverse trajectory data for training.
\end{itemize} \\
\cline{2-5}
&~\cite{VAEValid2023Se} & 
VAE within the constrained optimization in learned latent space (COIL) approach are used to optimize autonomous robot timings for deliveries, ensuring that an appropriate number of robots run simultaneously to maintain safety.  &
\begin{itemize}[leftmargin=3pt,rightmargin=0pt,after=\vspace{-1\baselineskip}]
    \item Achieves fast optimization and adjusts to delivery demands.
\end{itemize} &
\begin{itemize}[leftmargin=3pt,rightmargin=0pt,after=\vspace{-1\baselineskip}]
    \item Initial learning phase can be time and resource-intensive.
\end{itemize} \\
\cline{2-5}
&~\cite{li2021grin} & 
 Generative Relation and Intention Network (GRIN), a conditional generative model trained using variational inference and learning is deployed for multi-agent trajectory prediction.  &
\begin{itemize}[leftmargin=3pt,rightmargin=0pt,after=\vspace{-1\baselineskip}]
    \item Models uncertainties of intentions and relations comprehensively.
\end{itemize} &
\begin{itemize}[leftmargin=3pt,rightmargin=0pt,after=\vspace{-1\baselineskip}]
    \item May require a large amount of training data.
\end{itemize} \\
\hline
\multirow{1}{*}{\textbf{Others}} &~\cite{FUERTES2023106085} & 
Transformer architecture combined with DRL are deployed to optimize routing for multiple cooperative UAVs.  &
\begin{itemize}[leftmargin=3pt,rightmargin=0pt,after=\vspace{-1\baselineskip}]
    \item Offers better performance and efficient parallel processing compared to transitional algorithms like Compass~\cite{KOBEAGA201842}.
\end{itemize} &
\begin{itemize}[leftmargin=3pt,rightmargin=0pt,after=\vspace{-1\baselineskip}]
    \item May face computational speed challenges in certain scenarios.
\end{itemize} \\
\hline
\end{tabular}
\label{Autonomytable}
\end{table*}
 Another filed that GAI is utilized in is the scene understanding or captioning. Such a method includes using CLIP Prefix for image captioning, translating visual content of UAV-captured images into accurate text descriptions for decision making in UV~\cite{drones7020114}. Another method is deploying the Generative Knowledge-Supported Transformer (GKST), which enhances feature representation and retrieval performance by fusing image information from different view of vehicles.~\cite{rs15092221}. An interesting aspect of these technologies is their ability to process and interpret complex visual inputs, providing a level of contextual understanding that closely resembles human perception. This capability is particularly beneficial in dynamic environments, where rapid and accurate interpretation of visual data is crucial for effective decision-making.

 In summary, the generative capabilities of GAI prove to be invaluable in the field of environmental perception for UVs. From enhancing image resolution to generating synthetic dataset, creating diverse testing environments, and advancing scene understanding, GAI stands as a cornerstone technology driving the evolution and efficiency of UVs in comprehending and interacting with their surroundings.

\subsection{Level of Autonomy}

Autonomy refers to the capability of a system to perform a task or decision-making without human interventions~\cite{bi2021overview}. The level of autonomy represents the ability of an UV can operate independently while solely relying on its onboard sensors, algorithms, and computational resources. In UV swarms, the level of autonomy depends on various factors such as the type and complexity of the task, the ability to plan and execute route~\cite{mejias2021embedded}. Table \ref{Autonomytable} illustrates how the integration of GAI is pivotal in advancing these autonomous capabilities.

In the realm of UV swarm cooperative strategies, applications of GAI are exemplified by the integration of Generative Adversarial Imitation Learning (GAIL) with multi-agent DRL. For instance, the authors in \cite{xiang2022flock} introduce a Multi-Agent PPO-based Generative Adversarial Imitation Learning (MAPPO-GAIL) algorithm which employs multi-agent proximal policy optimization to sample trajectories concurrently, refining policy and value models. This algorithm incorporates grid probability for environmental target representation, increasing average target discovery probability by 73.33\% while only compromising 1.11\% average damage probability compared to traditional DRL search algorithms. Additionally, GAIL is utilized for training UAVs in virtual environments for navigation tasks, enabling adaptation to complex and dynamic scenes \cite{dronegial2023}.  

Furthermore, a VAE based model named the BézierVAE is proposed for modeling vehicle trajectories, particularly for safety validation. BézierVAE encodes trajectories into a latent space and decodes them using Bézier curves, thereby generating diverse trajectories. BézierVAE demonstrates a remarkable reduction in reconstruction errors of 91.3\% and unsmoothness by 83.4\% compared to traditional model TrajVAE~\cite{8569971}, significantly enhancing the safety validation of automated vehicles \cite{BézierVAE2019safetytrj}. For autonomous robot scheduling, COIL employs VAEs to generate optimized timing schedules, improving operational efficiency significantly~\cite{VAEValid2023Se}. Lastly, in multi-agent trajectory prediction, the GRIN model which is inspired by conditional VAE is adopted to forecast agent trajectories considering the complexities of intentions and social relations. Although complex systems face challenges such as adhering to contextual rules such as physical laws, challenges could be addressed by using a specific decoder or a surrogate model to approximate these limitations~\cite{li2021grin}. 

In route planning for UVs, transformer architecture combined with DRL is deployed to optimize routing for multiple cooperative UAVs. This method offers superior performance and efficient parallel processing, consistently achieving high rewards compared to traditional algorithms \cite{FUERTES2023106085}.

Enhancing autonomy in UVs is crucial for their independent and cooperative swarm operation. GAI's generative capabilities are applied in multiple aspects, from generating new trajectories to refining routing strategies and imitating expert agent routing behaviors in diverse scenarios. These diverse applications demonstrate dynamic and adaptable solutions crucial for UVs to efficiently and independently navigate and operate in complex and changing environments.

\subsection{Task/Resource Allocation}

\begin{table*}[!ht]
\renewcommand{\multirowsetup}{\centering}
\caption{Generative AI in Task/Resource Allocation}
\vspace*{2mm}
\footnotesize
\centering
\renewcommand\arraystretch{1.27}
\begin{tabular}{c||c<{\centering}||m{0.375\textwidth}|m{0.175\textwidth}|m{0.175\textwidth}}
\hline
\hline
\textbf{GAI Type} & \textbf{Reference} & \textbf{Description} & \textbf{Pros} & \textbf{Cons} \\
\hline
\hline
\multirow{2}{*}{\textbf{GAN}}  &\cite{9644885}  & GAIL-based algorithm is proposed to reconstruct a virtual environment for DRL where the generator learns to produce expert trajectories and the discriminator distinguish the expert trajectories from the generated trajectories.&
\begin{itemize}[leftmargin=3pt,rightmargin=0pt,after=\vspace{-1\baselineskip}]
    \item Mimics real-world environments for effective DRL.
    \item Faster learning and better performance.
\end{itemize} &
\begin{itemize}[leftmargin=3pt,rightmargin=0pt,after=\vspace{-1\baselineskip}]
    \item Real-world training may affect service quality.
\end{itemize} \\
\hline
\multirow{1}{*}{\textbf{VAE}} &\cite{RATHOD2023102133} &An autoencoder is applied to mitigate the issue of information ambiguity in the Hungarian algorithm. Specifically, the autoencoder reconstructs the data rate matrix when there is the same weight of the cellular user (CU) and Device-to-device user (D2DU) pair in the Hungarian algorithm. It provides an optimal reconstructed cost matrix using latent space as a hyperparameter.&
\begin{itemize}[leftmargin=3pt,rightmargin=0pt,after=\vspace{-1\baselineskip}]
    \item Provides optimal reconstructed cost matrix.
    \item Efficient resource allocation.
\end{itemize} &
\begin{itemize}[leftmargin=3pt,rightmargin=0pt,after=\vspace{-1\baselineskip}]
    \item Relies on simulated training data.
\end{itemize} \\
\hline
\multirow{3}{*}{\textbf{Diffusion}} &\cite{du2023task} & A diffusion model-based AI-generated optimal decision (AGOD) algorithm is proposed to address the optimal AIGC service provider (ASP) selection decision problem for the AIGC-as-a-Service architecture. &
\begin{itemize}[leftmargin=3pt,rightmargin=0pt,after=\vspace{-1\baselineskip}]
    \item Generates optimal selection decisions.
    \item Applicable to various optimization problems.
\end{itemize} &
\begin{itemize}[leftmargin=3pt,rightmargin=0pt,after=\vspace{-1\baselineskip}]
    \item Requires large-scale architectures.
    \item High resource consumption.
\end{itemize} \\
\cline{2-5}
&\cite{du2023yolobased} & A diffusion-based model is deployed to perform multi-step denoising on Gaussian noise and generates an optimal allocation scheme that maximizes the transmission quality of semantic information. &
\begin{itemize}[leftmargin=3pt,rightmargin=0pt,after=\vspace{-1\baselineskip}]
    \item Achieves higher transmission quality scores.
    \item Manages complex decision spaces.
\end{itemize} &
\begin{itemize}[leftmargin=3pt,rightmargin=0pt,after=\vspace{-1\baselineskip}]
    \item Requires extensive training.
\end{itemize} \\
\hline
\multirow{1}{*}{\textbf{Others}}&\cite{zou2023wireless}  &LLM are deployed for autonomously generating, executing, and prioritizing tasks in real-time, especially when deployed on-device, allowing for collaborative planning and problem-solving in wireless networks. &
\begin{itemize}[leftmargin=3pt,rightmargin=0pt,after=\vspace{-1\baselineskip}]
    \item Adapts to tasks with on-device intelligence.
\end{itemize} &
\begin{itemize}[leftmargin=3pt,rightmargin=0pt,after=\vspace{-1\baselineskip}]
    \item May produce errors due to network changes.
\end{itemize} \\
\hline
\end{tabular}
\label{tasktable}
\end{table*}

In the field of task and resource allocation for multi-agent UV swarms, GAI introduces effective approaches that enhance efficiency and adaptability of these systems. Traditional methods often rely on fixed algorithms and heuristic approaches, which may not always be sufficient for dynamic and complex environments \cite{skaltsis2021survey}. As illustrated in Table \ref{tasktable}, GAI offers the flexibility necessary for these challenging scenarios.

A GAIL-based algorithm is proposed to reconstruct virtual environments for DRL, where the generator produces expert trajectories and the discriminator distinguishes them from generated trajectories \cite{9644885}. This approach can make a virtual edge computing environment that closely mimics real-world conditions. It provides a place for computing resource allocation multi-agent DRL method to explore and infer reward function while avoiding impairing user experiences caused by arbitrary exploration. Furthermore, an autoencoder based method is applied to the Hungarian algorithm to mitigate information ambiguity issues caused by the same weight that appears in the data rate matrix, particularly in the allocation of bandwidth and power resources between cellular users (CU) and device-to-device users (D2DU)~\cite{RATHOD2023102133}. This method provides an optimal reconstructed cost matrix using latent space as a hyperparameter to assist in resource allocation decisions.

Additionally, the authors in \cite{du2023task} proposed a diffusion model-based AI-generated optimal decision (AGOD) algorithm. This algorithm enables adaptive and responsive task allocation based on real-time environmental changes and user demands. The algorithm's efficacy is further enhanced by the integration of DRL, as demonstrated in the Deep Diffusion Soft Actor-Critic (D2SAC) algorithm. Compared to traditional SAC methods, the D2SAC algorithm shows a performance improvement of approximately 2.3\% in terms of the task completion rate and 5.15\% in terms of utility gained \cite{du2023task}. Unlike traditional task allocation methods, which assume that all tasks and their corresponding utility values are known in advance, D2SAC could address the selection of the most appropriate service provider, where tasks arrive dynamically and in real time. D2SAC shows a notable performance improvement in terms of completion rate and utility gained compared to traditional methods.

In the realm of joint computing and communication resource allocation, the importance of effective management is accentuated in UVs due to their standalone nature and battery constraints. A diffusion-based model presented in \cite{du2023yolobased} offers an advanced method to design optimal energy allocation strategies for the transmission of semantic information. A key strength of this model is its ability to iteratively refine power allocation, ensuring transmission quality is optimized under varying conditions caused by the dynamic environment of UV swarm. With the transmission distance at 20 m and the transmission power at 4 kW, this diffusion model-based AI-generated scheme surpasses other traditional transmission power allocation methods like average allocation~(named Avg-SemCom) and Confidence-based Semantic Communication~(Conf-SemCom)~\cite{du2023yolobased} at approximately 500 iterations with 0.25 increase on transmission quality.

On the other hand, the authors in \cite{zou2023wireless}  proposed the incorporation of LLM explored to elevate the capabilities of GAI in task and resource allocation within multi-agent UV swarms. Using LLM's advanced decision-making and analysis ability, independent LLM instances for each user are created to breakdown the original intent ``reducing network energy consumption by $\Delta p = 0.85W$" into a series of detail tasks such as tuning transmit power and channel measurement. The results are then prompted to the LLM, which will add subsequent tasks and instruct related executors to take actions. With the integration on LLM, the UAV agents managed to achieve the power saving target in 2 rounds. Although further simulation results show that current GPT-4 experiences some difficulties in maintaining multiple goals when the number of agents increases. This integration signifies a significantly advancement in the autonomy and functionality of UV swarms.

In conclusion, GAI substantially advances the field of task and resource allocation in multi-agent UV swarms. From creating vivid simulation environments for the allocation algorithm to explore on, to iteratively adjust the allocation strategy and break a rough intent to detail tasks, GAI demonstrates a strong ability to handle dynamic environments and various challenges.

\subsection{Network Coverage and Peer-to-Peer Communication}

  \begin{table*}[!t]
\renewcommand{\multirowsetup}{\centering}
\caption{Generative AI in Network Coverage and Peer-to-Peer Communication}
\footnotesize
\centering
\renewcommand\arraystretch{1.27}
\begin{tabular}{c||c<{\centering}||m{0.375\textwidth}|m{0.175\textwidth}|m{0.175\textwidth}}
\hline
\hline
\textbf{GAI Type} & \textbf{Reference} & \textbf{Description} & \textbf{Pros} & \textbf{Cons} \\
\hline
\hline
\multirow{2}{*}{\textbf{GAN}}  &\cite{Marek2022Network}  & A cGAN is deployed to solve optimize network coverage by employing three pivotal elements: A generator that models and predicts optimal network configurations. A discriminator that evaluates the authenticity and efficiency of these configurations against real-world scenarios. And a unique encoding mechanism that maps these configurations into a latent space, ensuring adaptability and scalability.&
\begin{itemize}[leftmargin=3pt,rightmargin=0pt,after=\vspace{-1\baselineskip}]
    \item Ensures UAV optimal positioning.
    \item Reduce computational complexity.
\end{itemize} &
\begin{itemize}[leftmargin=3pt,rightmargin=0pt,after=\vspace{-1\baselineskip}]
    \item Require larger amount of training data.
\end{itemize} \\
\cline{2-5}
&~\cite{duan2023scma} &A network-assisted decoder balanced network and an autoencoder-based generative contradiction network (SCMA-TPGAN), which integrate a transformer as generator and PatchGAN as the discriminator, are utilized to optimize sparse code multiple access (SCMA) encoding and decoding, improving the bit error rate in uplink Rayleigh fading channels.&
\begin{itemize}[leftmargin=3pt,rightmargin=0pt,after=\vspace{-1\baselineskip}]
    \item Lower bit error rate.
    \item Enhance robustness against noise.
\end{itemize} &
\begin{itemize}[leftmargin=3pt,rightmargin=0pt,after=\vspace{-1\baselineskip}]
    \item Relies on base station's knowledge of the channel matrix and the codebook of each user.
\end{itemize} \\
\hline
\multirow{1}{*}{\textbf{Diffusion}} &\cite{du2023beyond} & GDMs are employed for image restoration and network optimization in vehicle-to-vehicle communication, allowing the restoration of transmitted images corrupted due to transmission disruptions and environmental noise.  &
\begin{itemize}[leftmargin=3pt,rightmargin=0pt,after=\vspace{-1\baselineskip}]
    \item Reduce data transfer and delay in communication
\end{itemize} &
\begin{itemize}[leftmargin=3pt,rightmargin=0pt,after=\vspace{-1\baselineskip}]
    \item Adds iterative computational load.
\end{itemize} \\
\hline
\multirow{1}{*}{\textbf{Others}}&\cite{9498892}&Self-attention-based transformer models are utilized to predict users' spatial distribution, guiding UAV Base Stations to adjust their positions for optimal network coverage.&
\begin{itemize}[leftmargin=3pt,rightmargin=0pt,after=\vspace{-1\baselineskip}]
    \item Outperforms LSTM in long sequences.
\end{itemize} &
\begin{itemize}[leftmargin=3pt,rightmargin=0pt,after=\vspace{-1\baselineskip}]
    \item May oversimplify real-world network dynamics.
\end{itemize} \\
\hline
\end{tabular}
\label{netwroktable}
\end{table*}
As mentioned in Section~\ref{USS}, a key application of UVs is their role as mobile base stations for the reconstruction of the communication network~\cite{akram2020multicriteria,li2019post,merwaday2016improved,malandrino2019multiservice,zhang2023sagin}. An effective positioning strategy is crucial in this context to ensure seamless access by achieving maximum user coverage with a limited number of UVs. Additionally, when UV swarms are deployed in hierarchical structures, with lead UVs acting as command centers, ensuring effective communication coverage among the sub-UVs is critical for task distribution and collaborations. As illustrated in Table \ref{netwroktable}, the need for efficient network coverage and vehicle-to-vehicle (V2V) communication is addressed by varieties of GAIs.

While utilizing UAVs as mobile stations to provide temporary network links in dynamic wireless communications is becoming increasingly popular, optimizing the network can be complex due to factors such as varying UAV altitudes, mobility patterns, spatial-domain interference distribution and external environmental conditions, which present unique challenges. In addressing the optimization of network coverage with limited UAVs, the authors in~\cite{Marek2022Network} propose the use of cGAN. This framework comprises a generator for modeling and predicting optimal network configurations, a discriminator for evaluating the efficiency of these configurations against real-world scenarios, and an encoding mechanism for adaptability and scalability. The method based on cGAN not only guarantees the best positioning of UAVs but also simplifies computational complexity, achieving $\mathcal{O}(k^2)$. In contrast, traditional methods like the core-set algorithm~\cite{badoiu2002approximate} and the spiral algorithm~\cite{lyu2016placement} have the time complexities of $\mathcal{O}(pk)$ and $\mathcal{O}(k^3)$, respectively where $p$ represents the number of user equipment in the area, while $k$ denotes the number of UAVs in the fleet~\cite{Marek2022Network}. Another solution proposed by the authors in  \cite{9498892} utilized the self-attention-based transformer to predict the user mobility and enhance the aerial base station placement. The transformer model was able to capture the spatio-temporal dependencies and handle long input and output sequences. The transformer-based scheme achieved significant gains in coverage rates compared to regular deployment schemes by improving the coverage rate by more than 31\% over the regular scheme~\cite{chaalal2020social} and by more than 9\% over the LSTM-based scheme.

In the domain of V2V communication, which is essential for secure navigation in the UV swarm, vehicles often relay images to communicate environmental data. However, these images can be corrupted due to transmission disruptions, environmental noise and noise caused by vehicle movement. To address this, the authors in \cite{du2023beyond} integrates GDMs for image restoration and network optimization. GDMs enable vehicles to restore transmitted images to their original quality by reducing data transfer and delay in communication. The iterative nature of GDMs, based on stochastic differential equations, is adept at refining internet-of-vehicles network solutions, notably in areas like path planning. For example, GDMs initiate optimization with a preliminary path, progressively enhancing it based on key performance indicators. The process capitalizes on these metric gradients to guide the path modifications towards an optimal solution. Compared to traditional DQN methods~\cite{mnih2013playing}, the proposed GDM-based method achieved a 100\% increase in average cumulative rewards at 300 epochs~\cite{du2023beyond}. 

In summary, for network coverage and accessibility, GAI can either generate positioning strategies directly or act as an encoder to enhance traditional algorithms by capturing spatial information. For efficiency, GAI acts as a framework that uses semantic information to reduce data transmission while maintaining communication by guided generation. However, while these developments represent a leap forward in managing UV swarms, there remain areas for further exploration. For example, the authors in~\cite{du2023beyond} open the question of integrating other modalities for more efficient communication. This suggests an opportunity for future research to investigate the incorporation of multimodal data processing in UV networks. Such explorations could significantly enhance the adaptability of these technologies to diverse network topology and environmental conditions. Additionally, the potential of GAI to facilitate autonomous decision-making within UV swarm deployments presents a promising avenue for advancing the field. By expanding the scope of GAI applications, researchers can further optimize UVs for a variety of complex real-world scenarios.

 \subsection{Security/Privacy}

     \begin{figure}[h!]
\centerline{\includegraphics[width=0.5\textwidth]{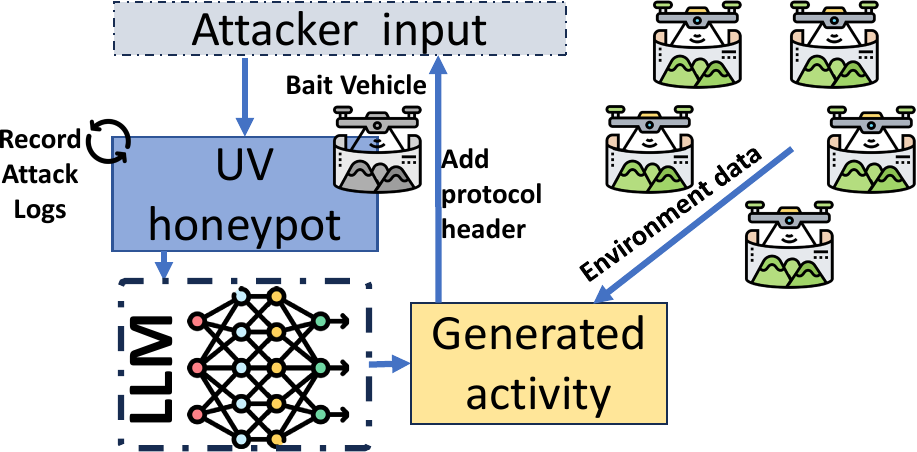}}
\caption{This framework highlights LLMs generating honeypot data, using Bait UV to pretend as a security hole. This tactic distracts attackers and logs their activities, enabling other UVs to bolster defenses against cyber threats. A strategic and collective deception that enhances overall swarm security.}
\label{llmhoneypot}
\end{figure}

 \begin{table*}[!t]
\renewcommand{\multirowsetup}{\centering}
\caption{Generative AI in Security/Privacy}
\footnotesize
\centering
\renewcommand\arraystretch{1.27}
\begin{tabular}{c||c<{\centering}||m{0.375\textwidth}|m{0.175\textwidth}|m{0.175\textwidth}}
\hline
\hline
\textbf{Type}&\textbf{Ref}&\textbf{Description}&\textbf{Pros}&\textbf{Cons} \\
\hline
\hline
\multirow{4}{*}{\textbf{GAN}} &\cite{xiong2019privacy}  &GAN-based image-to-image translation method named Auto-Driving GAN (ADGAN) is proposed to protect the location privacy of vehicular camera data by removing or modifying background buildings in images, while preserving the recognition utility of other objects such as traffic signs and pedestrians. &
\begin{itemize}[leftmargin=3pt,rightmargin=0pt,after=\vspace{-1\baselineskip}]
    \item Protect privacy while maintaining recognition accuracy.
    \item Enhance synthesis with multi-discriminator design.
\end{itemize} &
\begin{itemize}[leftmargin=3pt,rightmargin=0pt,after=\vspace{-1\baselineskip}]
    \item Potential privacy leaks via indirect data.
    \item May impair auto-driving quality and safety.
\end{itemize} \\
\cline{2-5}
&~\cite{liu2018trajgans} &TrajGANs is proposed protect the privacy of trajectory data by generating synthetic trajectories that follow the same distribution as the real data, while obscuring the individual locations and identities of users. &
\begin{itemize}[leftmargin=3pt,rightmargin=0pt,after=\vspace{-1\baselineskip}]
    \item Maintains data utility and pattern accuracy.
    \item Flexible in various data scenarios.
\end{itemize} &
\begin{itemize}[leftmargin=3pt,rightmargin=0pt,after=\vspace{-1\baselineskip}]
    \item Struggles with dense trajectory representation.
    \item Risk of overfitting and missing rare events.
\end{itemize} \\
\cline{2-5}
&~\cite{Rao2020LSTMTrajGANAD} &LSTM-TrajGAN generates synthetic trajectories that follow the same distribution as the real trajectories, but obscure the individual locations and identities of users. The synthetic trajectories can prevent users from being reidentified by the Trajectory-User Linking (TUL) algorithm, which links trajectories to users based on their spatial, temporal, and thematic characteristics thus expose vehicle privacy. &
\begin{itemize}[leftmargin=3pt,rightmargin=0pt,after=\vspace{-1\baselineskip}]
    \item Retains mobility patterns and data utility.
    \item Outperforms geomasking methods.
\end{itemize} &
\begin{itemize}[leftmargin=3pt,rightmargin=0pt,after=\vspace{-1\baselineskip}]
    \item Validation needed to prevent overfitting.
    \item Must weigh privacy against data utility.
\end{itemize} \\
\cline{2-5}
&~\cite{uittenbogaard2019privacy} &A framework using GAN is proposed to automatically segments and replaces moving objects (such as pedestrians and vehicles) from street-view images by inserting a realistic background.&
\begin{itemize}[leftmargin=3pt,rightmargin=0pt,after=\vspace{-1\baselineskip}]
    \item Segments and anonymizes moving objects.
\end{itemize} &
\begin{itemize}[leftmargin=3pt,rightmargin=0pt,after=\vspace{-1\baselineskip}]
    \item Risk of removing relevant static objects.
\end{itemize} \\
\hline
\multirow{1}{*}{\textbf{VAE}}&\cite{adeboye2022deepclean} &VAE is deployed to generate synthetic trajectories that randomize the released vehicle locations based on differential privacy, which adds noise to the data to prevent the disclosure of individual information.  &
\begin{itemize}[leftmargin=3pt,rightmargin=0pt,after=\vspace{-1\baselineskip}]
    \item Creates varied and realistic synthetic data.
    \item Adjustable privacy level.
\end{itemize} &
\begin{itemize}[leftmargin=3pt,rightmargin=0pt,after=\vspace{-1\baselineskip}]
    \item Introduce noise that may reduce data accuracy.
\end{itemize} \\
\hline
\multirow{2}{*}{\textbf{Others}}  &\cite{9857660}  &Federated Vehicular transformers combines federated learning with transformers for privacy-preserving computing and cooperation in autonomous driving. Transformers could achieve unified representation and fusion of multi-modal data, such as trajectories, images, and point clouds, from different vehicles  &
\begin{itemize}[leftmargin=3pt,rightmargin=0pt,after=\vspace{-1\baselineskip}]
    \item Privacy-centric performance balance.
\end{itemize} &
\begin{itemize}[leftmargin=3pt,rightmargin=0pt,after=\vspace{-1\baselineskip}]
    \item More complex system requirements.
\end{itemize} \\
\cline{2-5}
&~\cite{10141599} &A multi-class intrusion detection system (IDS) based on a transformer-based attention network for an in-vehicle controller area network (CAN) bus is proposed. The IDS can learn the features and correlations of CAN messages and classify them into different attack types using self-attention mechanisms.&
\begin{itemize}[leftmargin=3pt,rightmargin=0pt,after=\vspace{-1\baselineskip}]
    \item Accurate detection of diverse attacks.
    \item Efficient without detailed message labels.
\end{itemize} &
\begin{itemize}[leftmargin=3pt,rightmargin=0pt,after=\vspace{-1\baselineskip}]
    \item Longer training period.
    \item Venerable to new, unknown attack types.
\end{itemize} \\

\hline
\end{tabular}
\label{privacytable}
\end{table*}
Security and privacy are critical aspects in UV swarms, especially in military and surveillance applications. The integration of GAI in these domains offers innovative solutions for enhancing system security and ensuring privacy. As illustrated in Fig.~\ref{llmhoneypot}, an interesting potential application is to utilize GAI's ability to generate fake data or simulate communication activities to act as a honeypot to mislead potential attackers and reinforce system security~\cite{provos2004virtual}. The LLM-generated honeypots serve as an additional protective layer, disseminating false information to confuse and trap attackers, thereby enhancing the collective security of the swarm. This innovative use of language processing technology within the swarm network exemplifies a new frontier in safeguarding autonomous vehicles from sophisticated cyber threats. The use of GAI in UV swarm security and privacy protection is elaborated in Table~\ref{privacytable}.

One notable application of GAI in the realm of privacy protection is the Auto-Driving GAN (ADGAN)~\cite{xiong2019privacy}. ADGAN is a GAN-based image-to-image translation method designed to protect the privacy of vehicle camera location data. ADGAN achieves this by removing or modifying background buildings in images while retaining the utility of recognizing other objects like traffic signs and pedestrians. The semantic communication acts as an effective means to enhancing the security of UV swarms, as it removes the background images that are lrrelevant to tasks. Additionally, ADGAN introduces the multi-discriminator setting that enhances image synthesis performance and offers a stronger privacy protection guarantee against more powerful attackers \cite{xiong2019privacy}. Another similar application is a GAN-based framework that protects identity privacy in street-view images by altering recognizable features, such as replacing moving objects with a realistic background \cite{uittenbogaard2019privacy}.

In terms of trajectory data privacy, TrajGANs are employed to protect the privacy of trajectory data by generating synthetic trajectories~\cite{liu2018trajgans}. These trajectories follow the same distribution as real data while obscuring individual locations and identities of users. They preserve real data's statistical properties and capture human mobility patterns. However, TrajGANs may face challenges in creating dense representations of trajectories, particularly for timestamps and road segments, and may fail to identify some rare or exceptional events in the data. To further enhance the protection, the authors in~\cite{Rao2020LSTMTrajGANAD} present the LSTM-TrajGAN framework. The framework consists of three parts: a generator that generates and predicts realistic trajectory configurations, a discriminator that compares these configurations with real data to validate their authenticity and utility, and a specialized encoding mechanism that utilized LSTM~\cite{hochreiter1997long} recurrent neural network to perform space-time embedding of trajectory data and its respective time stamp. Its privacy protection efficacy is evaluated using a Trajectory-User Linking (TUL) algorithm as attackers~\cite{gao2017identifying}. Evaluated on a real-world semantic trajectory dataset, the proposed approach achieves better privacy protection by reducing the attacker's accuracy from 99.8\% to 45.9\% compared to traditional geo-masking methods like random perturbation at 66.8\% and Gaussian geo-masking at 48.6\%~\cite{gao2019exploring}. These results show that the LSTM-TrajGAN can better prevent users from being re-identified while preserving essential spatial and temporal characteristics of the real trajectory data. 

VAEs are also deployed for protecting UV trajectory privacy. The authors in \cite{adeboye2022deepclean} utilize a VAE to create synthetic vehicle trajectories, ensuring differential privacy by adding noise to data. This approach helps to effectively obscure vehicle locations, although some data distortions may be introduced due to the added noise. Transformers in federated learning, as discussed in \cite{9857660}, enhance privacy in autonomous driving by sharing only essential data features across networks. This method improves privacy but faces challenges with communication link stability and external interference.

To protect vehicle network security, the authors in \cite{10141599} proposed a transformer-based intrusion detection system that provides a sophisticated solution for vehicle networks. This system employs self-attention mechanisms to analyze Controller Area Network (CAN) messages, accurately classifying them into various in-vehicle attacks like denial-of-service, spoofing, and replay attacks. Another transformer-based model proposed by the authors in \cite{9857660} is the integration of transformers in federated learning setups. This method enables the sharing of key data features rather than raw data across a network of autonomous vehicles. This method significantly boosts privacy by minimizing the exposure of sensitive data while still enabling collaborative decision-making and computing.

In summary, the application of GAI in UV swarms has revolutionized security and privacy measures, particularly in sensitive sectors such as military and surveillance. Techniques like honeypots and GAN-based frameworks demonstrate GAI's capability in data manipulation for enhanced security. Additionally, the implementation of VAEs and transformers in federated learning for trajectory privacy, and advanced intrusion detection systems underscore GAI's adaptability and effectiveness in safeguarding against sophisticated cyber threats.

\subsection{Vehicle Safety and Fault Detection}

 \begin{table*}[!t]
\renewcommand{\multirowsetup}{\centering}
\caption{Generative AI in Vehicle Safety and Fault Detection}
\footnotesize
\centering
\renewcommand\arraystretch{1.27}
\begin{tabular}{c||c<{\centering}||m{0.37\textwidth}|m{0.17\textwidth}|m{0.185\textwidth}}
\hline
\hline
\textbf{GAI Type} & \textbf{Reference} & \textbf{Description} & \textbf{Pros} & \textbf{Cons} \\
\hline
\hline
\multirow{1}{*}{\textbf{GAN + VAE}}  &\cite{Marek2022Network}  &  VAE-CGAN is deployed to complements missing time-series data in case of sensor faults, where VAE helps in initializing and correcting the noise in traditional CGAN, and CGAN’s discriminator improves the sample quality generated by VAE’s decoder.&
\begin{itemize}[leftmargin=3pt,rightmargin=0pt,after=\vspace{-1\baselineskip}]
    \item High-quality samples even with different missing rates.
    \item Accurate predictions.
\end{itemize} &
\begin{itemize}[leftmargin=3pt,rightmargin=0pt,after=\vspace{-1\baselineskip}]
    \item Require longer training period.
\end{itemize} \\
\hline
\multirow{1}{*}{\textbf{VAE}} 
&~\cite{dhakal2023uav} &VAE is deployed learn to encode normal UAV data into a lower-dimensional space then reconstruct it back and identify anomalies by comparing the reconstruction error with a deviation threshold.&
\begin{itemize}[leftmargin=3pt,rightmargin=0pt,after=\vspace{-1\baselineskip}]
    \item Able to detect uncovering novel faults.
    \item Accurate predictions.
\end{itemize} &
\begin{itemize}[leftmargin=3pt,rightmargin=0pt,after=\vspace{-1\baselineskip}]
    \item Offline operation. 
    \item Lack of generalizability.
\end{itemize} \\
\hline
\multirow{2}{*}{\textbf{Others}}  &\cite{sadhu2020board,sadhu2023board}  & LSTM network is utilized to detect faults in UVs by analyzing sensor data to identify patterns indicative of anomalies, and it operates by processing data in a sequence to capture temporal dependencies that are crucial for recognizing irregularities in the sensor readings. &
\begin{itemize}[leftmargin=3pt,rightmargin=0pt,after=\vspace{-1\baselineskip}]
    \item Accurate predictions.
\end{itemize} &
\begin{itemize}[leftmargin=3pt,rightmargin=0pt,after=\vspace{-1\baselineskip}]
    \item Require a large amount of data for training.
    \item Computationally intensive.
\end{itemize} \\
\cline{2-5}
&~\cite{zhao2023battery} &BERT based mode is deployed for extracting features across multiple spatio-temporal scales to identifies five common anomaly types for battery fault diagnosis.&
\begin{itemize}[leftmargin=3pt,rightmargin=0pt,after=\vspace{-1\baselineskip}]
    \item Reliable predictions of battery faults using onboard sensor data.
    \item Offers significant lead time for preventive measures.
\end{itemize} &
\begin{itemize}[leftmargin=3pt,rightmargin=0pt,after=\vspace{-1\baselineskip}]
    \item Computationally intensive.
    \item Risk of misidentifying battery conditions.
\end{itemize} \\
\hline
\end{tabular}
\label{SAFETYtable}
\end{table*}
Vehicle safety is another critical concern that encompasses the detection, isolation, and resolution of system faults. Unlike other safety concerns such as collision avoidance or the development of safe path planning strategies for UV swarms, which are more closely related to the level of autonomy of these systems~\cite{9981803}, research on UV safety highlights the unique challenges brought forth by internal vulnerabilities of UV systems, including algorithmic and hardware failures. Research in this field aims to enhance the overall reliability and safety of UV operations by developing methodologies and technologies that enable these systems to effectively identify and rectify potential faults before they impact vehicle performance or safety.

Monitoring operational parameters for fault detection in UV systems is essential to ensure their safety and efficiency. A novel framework has been proposed that uses LSTM networks combined with autoencoders, enabling continuous learning from vehicle performance data~\cite{sadhu2020board}. This framework enhances the system's ability to pinpoint and address faults progressively. The ability of LSTM in handling time-series data makes this approach particularly effective in dynamic environments where various factors can influence vehicle performance.The LSTM autoencoder can generate synthetic data points that represent potential fault scenarios. This enhances the training dataset, allowing the model to learn from a wider range of conditions and achieve 90\% accuracy in detecting and 99\% accuracy in classifying different types of drone misoperations based on simulation data. This significantly improves the safety and operational efficiency of UV systems. In subsequent developments~\cite{sadhu2023board}, the advances in UAV fault detection and classification, particularly the four-fold increase in speed through FPGA-based hardware acceleration while half the energy consumption. This research further identifies a key consideration for GAI, which shows that model computation can be optimized for real-time operations. The successful deployment in UAV swarms also suggests that similar strategies could enhance GAI performance in dynamic environments and complex task coordination.

On the other hand, VAE presents a sophisticated approach to fault and anomaly detection in UV swarms. The authors in~\cite{dhakal2023uav} proposed a new method by training VAE on data that represents the UVs' normal operations. This method helps VAEs develop an understanding of what constitutes standard performance. The learning process involves the reconstruction of input data, where the model's ability to accurately replicate the original data serves as the basis for identifying operational consistency. A significant deviation in the reconstruction error from the norm signals a potential fault or anomaly. By generating reconstructions of the input data and calculating the resulting errors, the VAE-based method achieved an average accuracy of 95.6\% in detecting faults and anomalies~\cite{dhakal2023uav}. The advantage of utilizing the VAEs ability of mapping relations is their proficiency in uncovering novel faults or issues that were not present or accounted for in the training dataset. This feature ensures that VAE-based systems can maintain high levels of safety and reliability in diverse and unpredictable scenarios. This feature is invaluable in the context of UV operations, which often encounter a broad spectrum of environmental conditions and operational challenges. Nonetheless, it is critical to acknowledge that the performance of VAEs can be affected by various factors that include the complexity of the VAE model itself, the quality and diversity of the data used for training, and the specific threshold set for flagging reconstruction errors as potential faults. 

Furthermore, the authors in~\cite{zhao2023battery} utilize a spatio-temporal Transformer network for battery fault diagnosis and failure prognosis in electric vehicles due to its specialized architecture, which performs well at extracting key features across multiple spatial and temporal scales. The adoption of a spatio-temporal Transformer network for battery fault diagnosis and failure prognosis in vehicles excels in identifying early warning signals and predicting faults over varying spatial and temporal scales. Its ability to analyze and predict the evolution of battery faults using onboard sensor data aligns perfectly with the needs of UV, which are heavily reliant on battery integrity for their operation. By integrating such a model, predictive maintenance strategies are greatly enhanced, allowing early detection of anomalies and prediction of battery failures within a precise window ranging from 24 hours to a week. This approach not only enhances operational efficiency by optimizing vehicle schedules to reduce downtime but also plays a crucial role in safeguarding against potential battery failures that could compromise vehicle safety.

In UV operations, ensuring safety and reliability not only involves detecting faults, but also isolating affected components to prevent further issues, and implementing targeted solutions for resolution. For instance, in the case of relatively minor issues such as the loss of information due to sensor malfunctions, the utilization of VAE and GAN illustrates the innovative application of GAI in fault management~\cite{ling2023optimization}. Through optimization of the VAE-CGAN structure, these models can regenerate missing time series data, demonstrating their effectiveness in scenarios where operational faults compromise data integrity. This function is especially beneficial for applications like UAV-based agricultural surveillance, where the continuity of data collection is paramount.

In tackling severe issues that compromise UV swarm operations, an intriguing aspect of current research is the development of strategies for ``where to crash" decision protocols that stand out~\cite{tong2022evaluation}. This concept addresses the need for predefined protocols on how and where a UV should terminate its operation in the event of a critical failure to minimize secondary hazards. These protocols range from emergency landing zones for UAVs to specific sinking points for USVs and UUVs, and controlled stop measures for UGVs. However, these predefined protocols may not be able to accommodate all possible scenarios. Thus, integrating GAI into UV swarm fault management strategies offers an advanced method to enhance safety. For example, by analyzing real-time sensor data and understanding the intricacies of swarm dynamics, Transformers are able to make context-aware decisions to identify the safest termination points for compromised UVs accurately~\cite{8342891}. Incorporating such GAI may not only improve the management of critical failures, but also lower the risk of secondary incidents.

\section{Open Issues and Future Research Directions}
\label{open_issue}
As discussed above, GAI has the great potential to further enhance UV swarms in various aspects. However, due to the complex and dynamic nature of UV swarms, several issues will need to be tackled. As such, this section aims to highlight open issues of GAI in UV swarms and their potential solutions.

\subsection{Scalability}

In the future, there can be a large number of UVs in a swarm to perform complex tasks in challenging environments such as precision agriculture, environmental monitoring, military operations, and delivery services. This introduces several issues to the development of GAI in UV swarms. Specifically, as the number of UVs increases, coordinating movements and communications between them become much more complex due to various factors like congestion, latency, signal interference, and limited communication range. This demands for novel GAI approaches that can quickly determine optimal movements and actions for each UV in such complex situations. One potential direction is to design distributed GAI architectures based on federated learning for collaborations between different groups of unmanned swarms, instead of relying on a single server/swarm leader. By doing this, the computing load can be distributed among swarm leaders, resulting in a more efficient learning and collaborating process between different groups of unmanned swarms, especially in large-scale swarm settings.

\subsection{Adaptive GAI}
In UV swarms, system conditions are highly dynamic and uncertain due to the mobility of UVs as well as surrounding complex environments. Although GAI has abilities to deal with these uncertainties, it is essential to develop adaptive GAI approaches to further reduce the system latency when retraining GAI models under new environmental conditions, especially in large-scale unmanned swarms. Integrating GAI with recent advances in AI such as transfer learning and meta-learning is a potential solution. In particular, transfer learning aims to transfer knowledge learned in a source environment to facilitate the learning process of similar tasks in new environments while meta-learning aims to learn how to learn to speed up the learning process in new conditions. These techniques can help AI models achieve good training accuracy with a few training samples from the new environments~\cite{pan2009survey, hospedales2021meta}. In addition, advanced deep reinforcement learning can also be integrated into GAI models for real-time learning and feedback on the behaviors of UV swarms.

\subsection{AI-native UV Networks}

In large-scale and heterogeneous UV networks, the energy efficiency, latency and security issues are the most critical for coordination locally and globally. 
In this operational circumstance, semantic communication (SemCom) with knowledge base (KB) shared among UVs is a viable solution to minimize the transmission overhead by eliminating the task-irrelevant information while performing a given task reliably, which greatly improves the demanding energy efficiency, latency and security. 
The UVs are limited in both the resource and hardware, and hence it is challenging to realize AI-native UV networks that enable the SemCom with shared KB. 
It should be noted that the on-going connections among internet of things (IoT), namely UVs are inherently task-oriented, dynamic and short-term due to their mobility, and achieving the above goal via the SemCom in such a circumstance is of paramount importance. 
The GAI approaches herein are considered to be effective in dealing with the dynamic and complex environments of such large-scale and heterogeneous UV networks, which are worthy of further research towards the AI-native UV networks.

\subsection{3-D Interference Control}

In large-scale and heterogeneous UV networks with high mobility, the resource-constrained multiple UV connections face with dynamic interference patterns which are distributed across 3-D space and temporal domain, and therefore a sophisticated interference control should be exercised for coordinating and maintaining communication among diverse UVs with limited power usage. Especially, the high mobility renders this interference control challenging in such complex environments, where the GAI approaches are proven to be effective in regulating the dynamic interference fluctuations within 3-D coverage. 

\subsection{Security and Privacy}

As reviewed above, GAI can help to improve the security and privacy of UVs by generating synthetic data to deceive attackers. Nevertheless, GAI, and AI in general, are vulnerable to adversarial attacks in which attackers try to disrupt the training process of AI models by poisoning training data, evading trained models, and extracting model information. More dangerously, attackers can also leverage GAI to generate ``fake'' data that is hard to distinguish. In addition, in UV swarms with resource-constrained vehicles, e.g., UAVs, individual vehicles are extremely vulnerable to adversarial attacks, and cyber attacks in general, due to the lack of computing and energy resources to perform sophisticated and efficient countermeasures. Dealing with adversarial attacks in GAI is still a new research problem and there are limited efforts in the literature. As such, there is an urgent demand for lightweight and effective solutions to improve the security of UV swarms. One potential solution is to leverage GAI to recover poisoned training data. Integrating deep reinforcement learning with human feedback can also be useful in defeating adversarial attacks. Finally, federated learning can also be adopted to enhance the privacy of unmanned swarms since data of vehicles (e.g., their sensing data and locations) will not be shared with others.

\section{Conclusion}
\label{conclusion}
In this paper, we have provided a comprehensive survey on the applications of GAI for UV swarms. We first provide an in-depth overview of UVs and UV swarms as well as their applications and existing challenges. Then, we present the fundamentals of various GAI techniques and their capabilities in addressing the challenges of UV swarms. After that, a comprehensive review on the applications of GAI for emerging problems in UV swarms is presented with insights and discussions. Finally, we discuss the opening issues of GAI in UV swarms and provide several potential research directions.
\bibliographystyle{IEEEtran}
\bibliography{citations}
\begin{IEEEbiography}[{\includegraphics[width=1in,height=1.25in, clip,keepaspectratio]{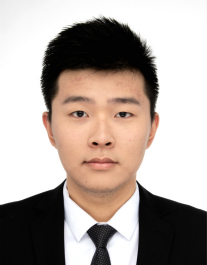}}]{Guangyuan Liu} is currently pursuing the Ph.D. degree with the School of Computer Science and Engineering, Energy Research Institute @ NTU, Nanyang Technological University, Singapore, under the Interdisciplinary Graduate Program. He received the B.Sc. degree from Nanyang Technological University, Singapore, in 2022. He won the Honorary Mention award in the ComSoc Student Competition from IEEE Communications Society in 2023, and the First Prizes in the 2024 ComSoc Social Network Technical Committee (SNTC) Student Competition. His research interests include generative AI, computer vision and resource allocation.
\end{IEEEbiography}
\vspace{-1cm}

\begin{IEEEbiography}[{\includegraphics[width=1in,height=1.25in, clip,keepaspectratio]{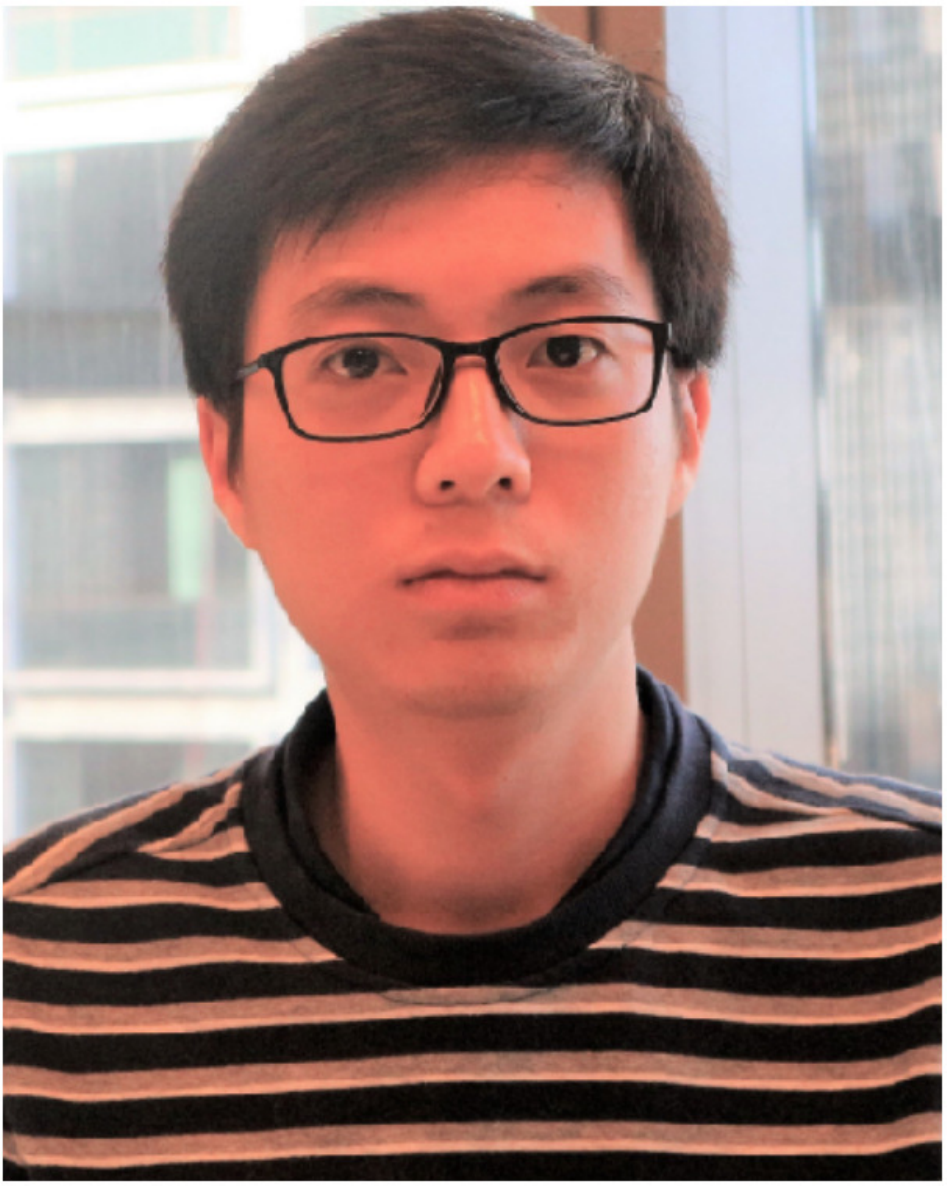}}]{Nguyen Van Huynh} received the B.E. degree in electronics and telecommunications engineering from the Hanoi University of Science and Technology, Vietnam, in 2016, and the Ph.D. degree from the University of Technology Sydney (UTS), Australia, in 2021. He is currently a Lecturer with the School of Computing, Engineering and the Built Environment, Edinburgh Napier University (ENU), U.K. Before joining ENU, he was a PostDoctoral Research Associate with the Department of Electrical and Electronic Engineering, Imperial College London, U.K. His research interests include cybersecurity, 5G/6G, the IoT, and machine learning.
\end{IEEEbiography}
\vspace{-1cm}
\begin{IEEEbiography}[{\includegraphics[width=1in,height=1.25in, clip,keepaspectratio]{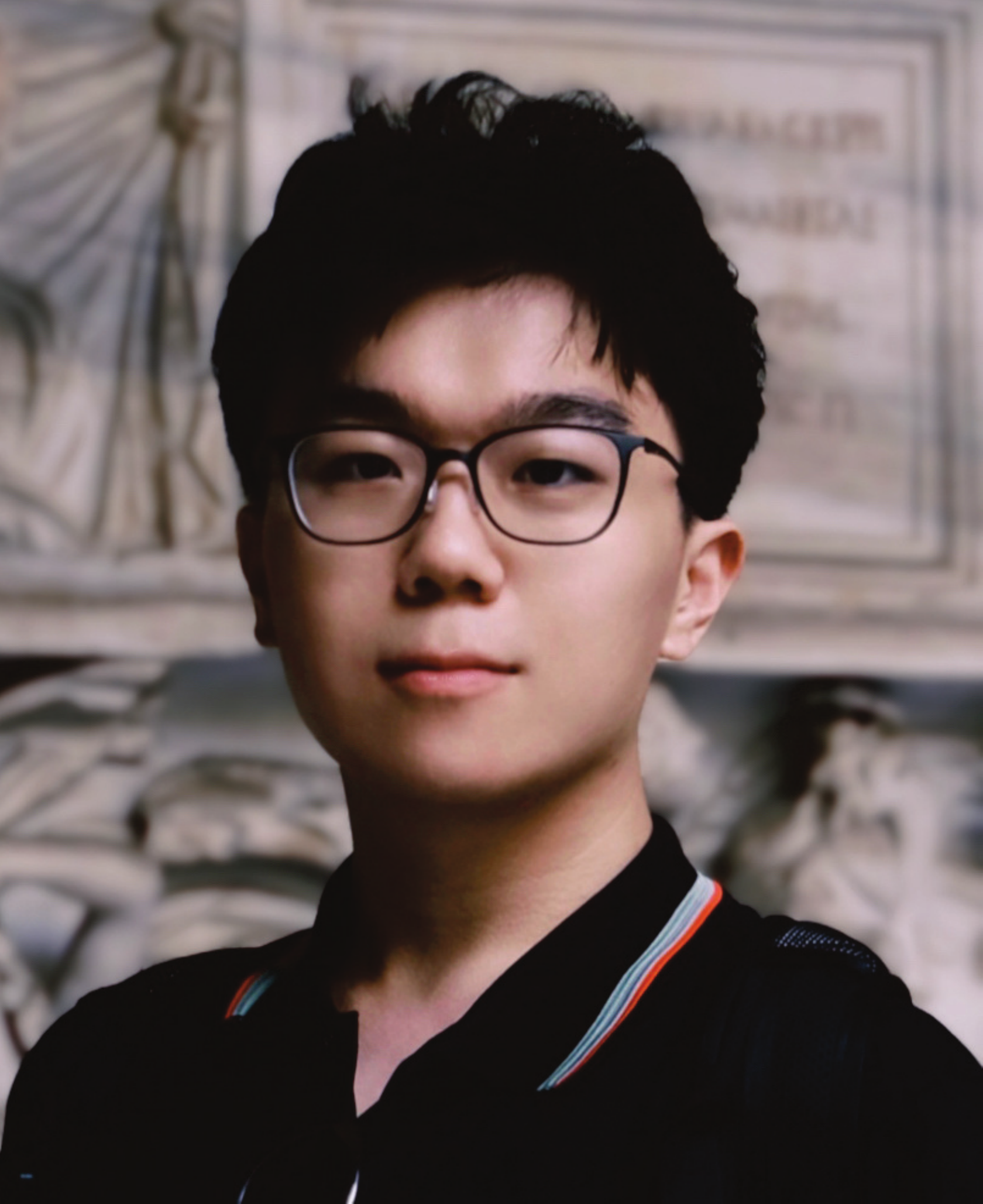}}]{Hongyang Du} is currently pursuing the Ph.D. degree with the School of Computer Science and Engineering, Energy Research Institute @ NTU, Nanyang Technological University, Singapore, under the Interdisciplinary Graduate Program. He received the B.Sc. degree from Beijing Jiaotong University, Beijing, China, in 2021. He is the Editor-in-Chief assistant of IEEE Communications Surveys \& Tutorials (2022-2024). He was recognized as an exemplary reviewer of the IEEE Transactions on Communications and IEEE Communications Letters in 2021. He was the recipient of the IEEE Daniel E. Noble Fellowship Award from the IEEE Vehicular Technology Society in 2022, the recipient of the IEEE Signal Processing Society Scholarship from the IEEE Signal Processing Society in 2023, the recipient of Chinese Government Award for Outstanding Students Abroad in 2023, and the recipient of the Singapore Data Science Consortium (SDSC) Dissertation Research Fellowship in 2023. He won the Honorary Mention award in the ComSoc Student Competition from IEEE Communications Society in 2023, and the First and Second Prizes in the 2024 ComSoc Social Network Technical Committee (SNTC) Student Competition. His research interests include semantic communications, generative AI, and resource allocation.
\end{IEEEbiography}
\vspace{-1cm}
\begin{IEEEbiography}[{\includegraphics[width=1in,height=1.25in, clip,keepaspectratio]{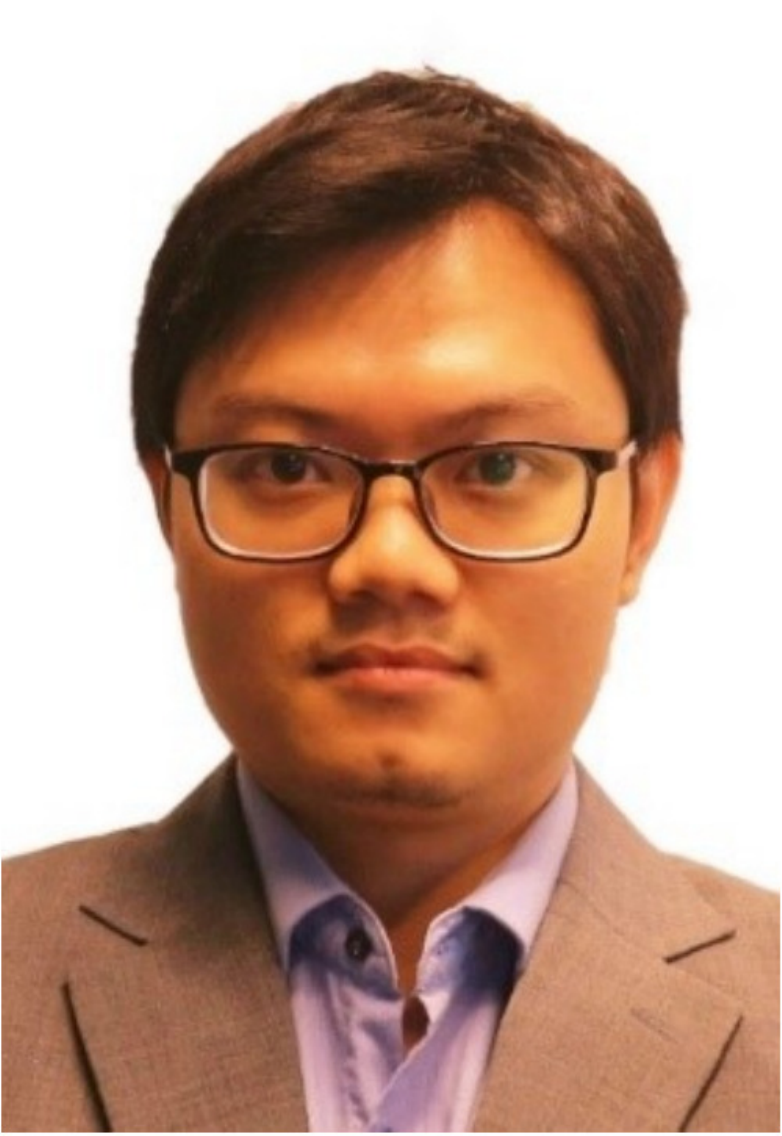}}]{Dinh Thai Hoang} is currently a faculty member at the School of Electrical and Data Engineering, University of Technology Sydney, Australia. He received his Ph.D. in Computer Science and Engineering from the Nanyang Technological University, Singapore, in 2016. His research interests include emerging topics in wireless communications and networking such as machine learning, ambient backscatter communications, IRS, edge intelligence, cybersecurity, IoT, and 5G/6G networks. He has received several awards including Australian Research Council and IEEE TCSC Award for Excellence in Scalable Computing (Early Career Researcher). Currently, he is an Editor of IEEE Transactions on Wireless Communications, IEEE Transactions on Cognitive Communications and Networking and Associate Editor of IEEE Communications Surveys \& Tutorials.
\end{IEEEbiography}
\newpage
\begin{IEEEbiography}[{\includegraphics[width=1in,height=1.25in, clip,keepaspectratio]{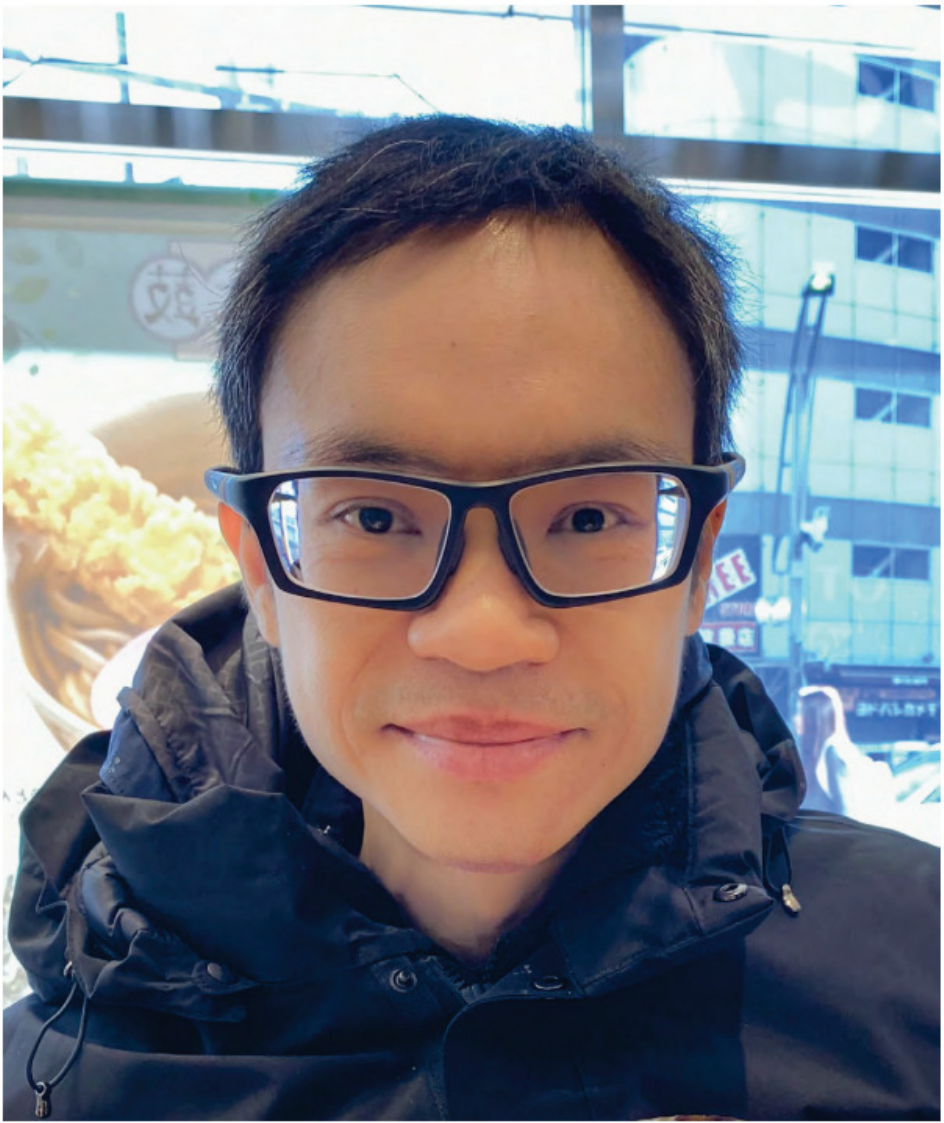}}]{Dusit Niyato} (Fellow, IEEE) is currently a professor in the School of Computer Science and Engineering and, by courtesy, School of Physical and Mathematical Sciences, at the Nanyang Tech nological University, Singapore. He received B.E. from King Mongkuk’s Institute of Technology Lad-krabang (KMITL), Thailand in 1999 and Ph.D. in Electrical and Computer Engineering from the University of Manitoba, Canada in 2008. He has authored four books including ``Game Theory in Wireless and Communication Networks: Theory, Models, and Applications" with Cambridge University Press. He won the Best Young Researcher Award of IEEE Communications Society (ComSoc) Asia Pacific (AP) and The 2011 IEEE Communications Society Fred W. Ellersick Prize Paper Award. Currently, he is serving as a senior editor of IEEE Wireless Communications Letter, an area editor of IEEE Transactions on Wireless Communications (Radio Management and Multiple Access), an area editor of IEEE Communications Surveys and Tutorials (Network and Service Management and Green Communication), an editor of IEEE Transactions on Communications, an associate editor of IEEE Transactions on Mobile Computing, IEEE Transactions on Vehicular Technology, and IEEE Transactions on Cognitive Communications and Networking.
\end{IEEEbiography}

\begin{IEEEbiography}[{\includegraphics[width=1in,height=1.25in,clip,keepaspectratio]{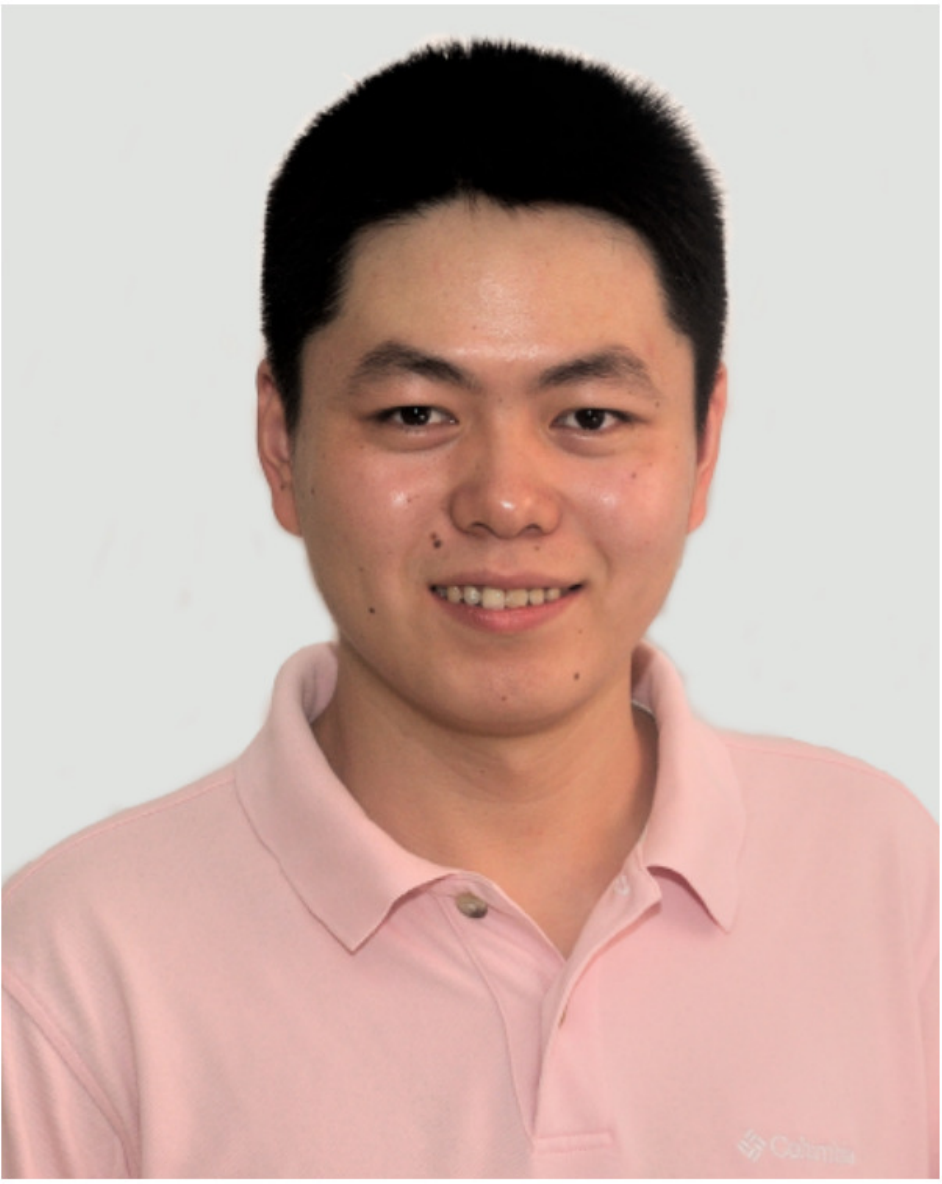}}]{Kun Zhu} received the Ph.D.degree from the School of Computer Engineering, Nanyang Technological University, Singapore, in 2012. He was a Research Fellow with the Wireless Communications, Networks, and Services Research Group, University of Manitoba, Canada. He is currently a Professor with the College of Computer Science and Technology, Nanjing University of Aeronautics and Astronautics, China. His research interests include resource allocation in 5G, wireless virtualization, and self-organizing networks. He has served as a TPC member for several conferences and a reviewer for several journals.
\end{IEEEbiography}

\begin{IEEEbiography}[{\includegraphics[width=1in,height=1.25in,clip,keepaspectratio]{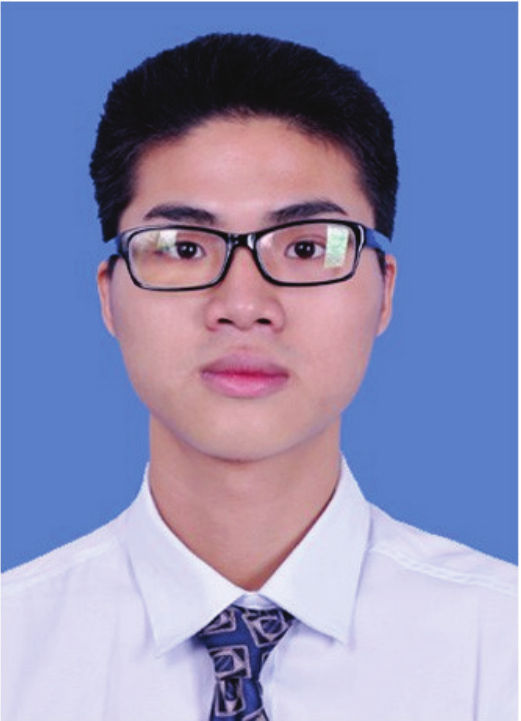}}]{Jiawen Kang} received the Ph.D. degree from the Guangdong University of Technology, China in 2018. He was a postdoc at Nanyang Technological University, Singapore from 2018 to 2021. He currently is a full professor at Guangdong University of Technology, China.  His research interests mainly focus on blockchain, security, and privacy protection in wireless communications and networking. He has published more than 160 research papers in leading journals and flagship conferences including 12 ESI highly cited papers and 3 ESI hot papers. He has won IEEE VTS Best Paper Award, IEEE Communications Society CSIM Technical Committee Best Journal Paper Award, IEEE Best Land Transportation Paper Award, IEEE HITC Award for Excellence in Hyper-Intelligence Systems (Early Career Researcher award), IEEE Computer Society Smart Computing Special Technical Community Early-Career Award, and 13 best paper awards of international conferences.
\end{IEEEbiography}

\begin{IEEEbiography}[{\includegraphics[width=1in,height=1.25in,clip,keepaspectratio]{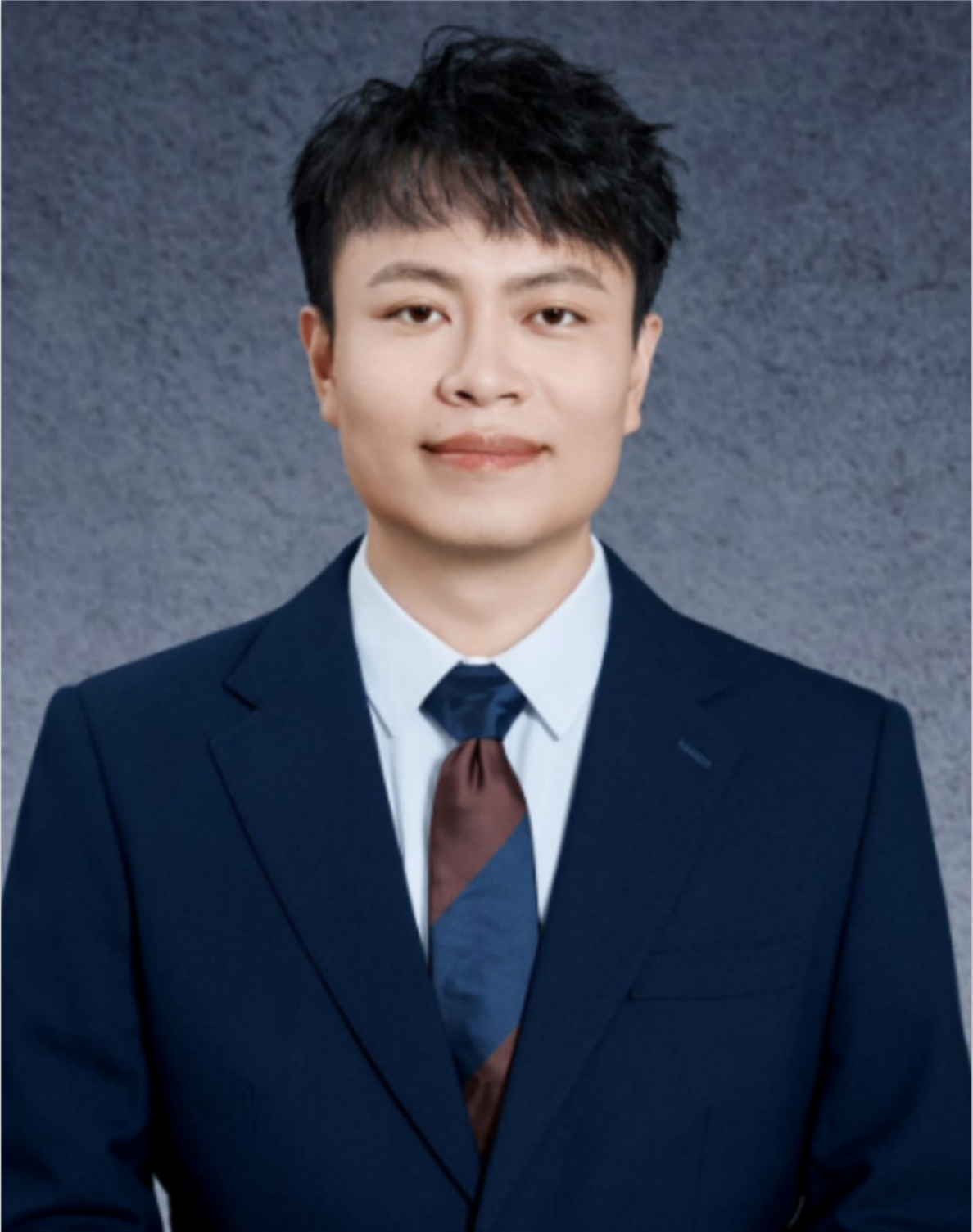}}]{Zehui~Xiong} is an Assistant Professor at Singapore University of Technology and Design (SUTD), and also an Honorary Adjunct Senior Research Scientist with Alibaba-NTU Singapore Joint Research Institute, Singapore. He obtained the B.Eng degree with the highest honors in Telecommunications Engineering at Huazhong University of Science and Technology (HUST), Wuhan, China. He received the Ph.D. degree in Computer Science and Engineering at Nanyang Technological University (NTU), Singapore. He was a visiting scholar with Department of Electrical Engineering at Princeton University and a visiting scholar with Broadband Communications Research (BBCR) Lab in Department of Electrical and Computer Engineering at University of Waterloo. His research interests include wireless communications, Internet of Things, blockchain, edge intelligence, and Metaverse.  Recognized as a Highly Cited Researcher, he has published more than 200 peer-reviewed research papers in leading journals and flagship conferences.
\end{IEEEbiography}

\begin{IEEEbiography}[{\includegraphics[width=1in,height=1.25in, clip,keepaspectratio]{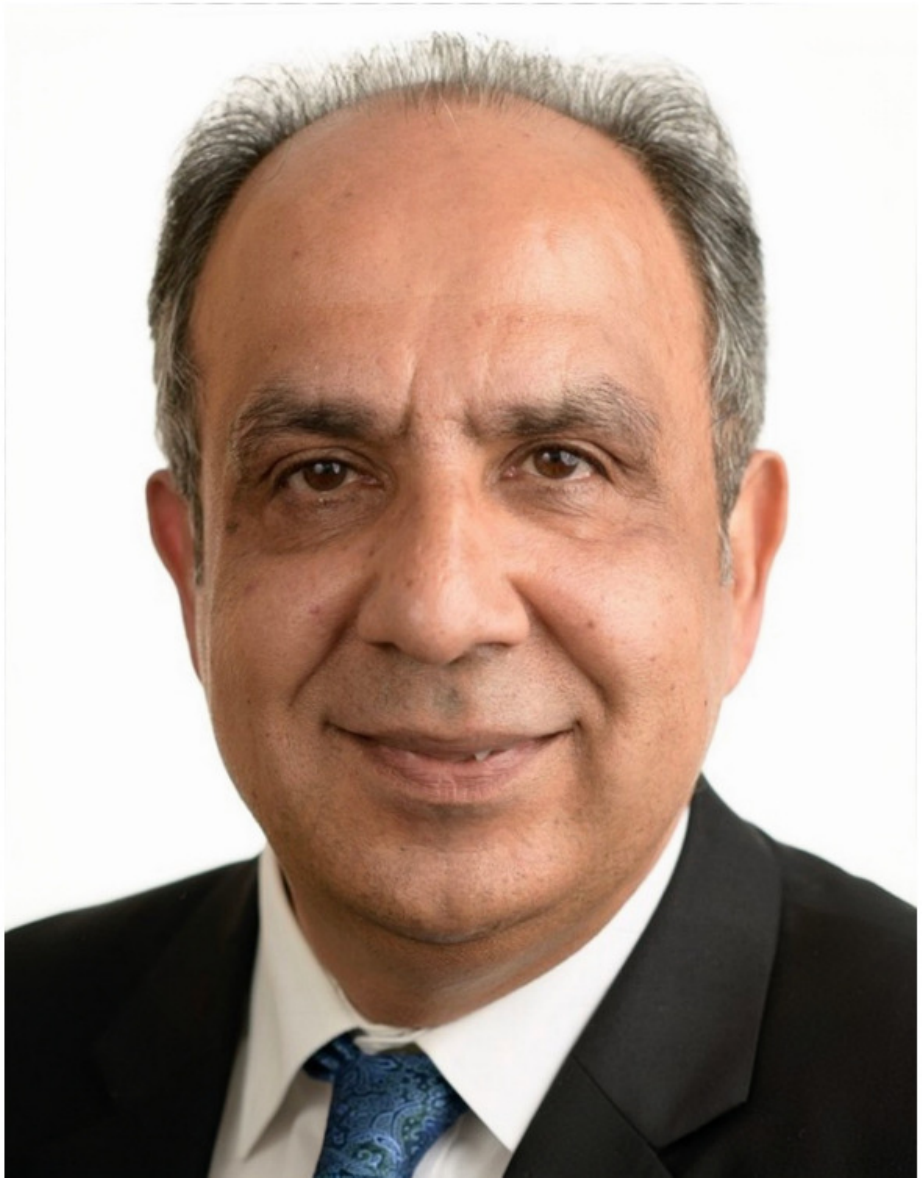}}]{Abbas Jamalipour} (Fellow, IEEE) received the Ph.D. degree in electrical engineering from Nagoya University, Nagoya, Japan, in 1996. He holds the position of a Professor of ubiquitous mobile networking with The University of Sydney. He has authored nine technical books, 11 book chapters, over 550 technical articles, and five patents, all in the area of wireless communications and networking. He is a fellow of the Institute of Electrical, Information, and Communication Engineers (IEICE) and the Institution of Engineers Australia, an ACM Professional Member, and an IEEE Distinguished Speaker. He was a recipient of several prestigious awards, such as the 2019 IEEE ComSoc Distinguished Technical Achievement Award in Green Communications, the 2016 IEEE ComSoc Distinguished Technical Achievement Award in Communications Switching and Routing, the 2010 IEEE ComSoc Harold Sobol Award, the 2006 IEEE ComSoc Best Tutorial Paper Award, and over 15 best paper awards. He has been the General Chair or the Technical Program Chair for several prestigious conferences, including IEEE ICC, GLOBECOM, WCNC, and PIMRC. He was the President of the IEEE Vehicular Technology Society from 2020 to 2021.
\end{IEEEbiography}

\begin{IEEEbiography}[{\includegraphics[width=1in,height=1.25in, clip,keepaspectratio]{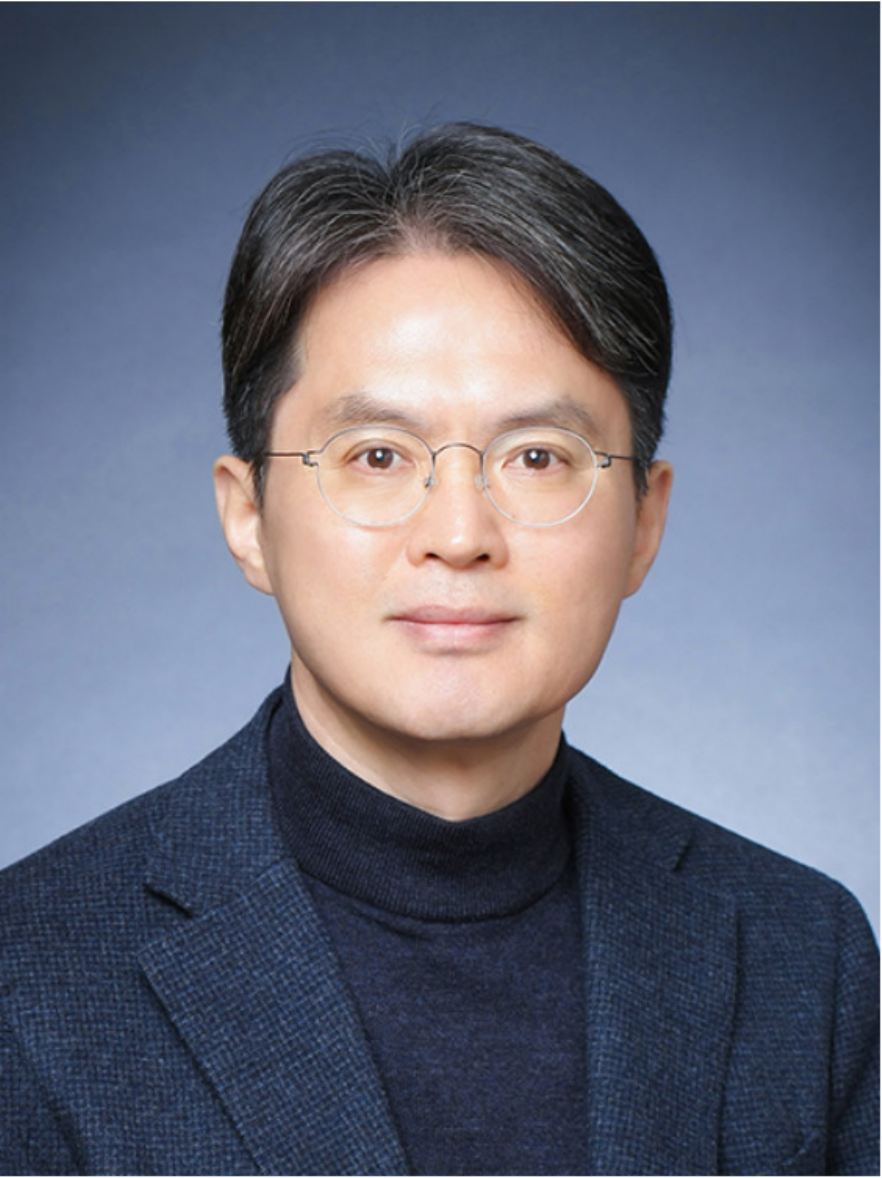}}]{Dong In Kim} (Fellow, IEEE) received the Ph.D. degree in electrical engineering from the University of Southern California, Los Angeles, CA, USA, in 1990. He was a Tenured Professor with the School of Engineering Science, Simon Fraser University, Burnaby, BC, Canada. He is currently a Distinguished Professor with the College of Information and Communication Engineering, Sungkyunkwan University, Suwon, South Korea. He is a Fellow of the Korean Academy of Science and Technology and a Member of the National Academy of Engineering of Korea. He was the first recipient of the NRF of Korea Engineering Research Center in Wireless Communications for RF Energy Harvesting from 2014 to 2021. He received several research awards, including the 2023 IEEE ComSoc Best Survey Paper Award and the 2022 IEEE Best Land Transportation Paper Award. He was selected the 2019 recipient of the IEEE ComSoc Joseph LoCicero Award for Exemplary Service to Publications. He was the General Chair of the IEEE ICC 2022, Seoul. Since 2001, he has been serving as an Editor, an Editor at Large, and an Area Editor of Wireless Communications I for IEEE Transactions on Communications.
\end{IEEEbiography}

\end{document}